\documentclass[twoside,11pt]{article}

\usepackage{blindtext}
\usepackage{graphicx}
\usepackage{caption}
\usepackage{subcaption}
\usepackage[preprint]{jmlr2e}

\newcommand{\fracpartial}[2]{\frac{\partial #1}{\partial  #2}}

\newcommand{\atone}[2][]{%
  \ensuremath{%
    \ifx\relax#1\relax
      \boldsymbol{\gamma}^{(#2)}%
    \else
      \gamma_{#1}^{(#2)}%
    \fi
  }%
}

\newcommand{\attwo}[2][]{%
    \ensuremath{%
      \ifx\relax#1\relax
        \boldsymbol{\tilde{\gamma}}^{(#2)}%
      \else
        \tilde{\gamma}_{#1}^{(#2)}%
      \fi
    }%
}

\newcommand{\qone}[2][]{%
    \ensuremath{%
        \boldsymbol{\theta}_{#1}^{(#2)}%
    }%
}

\newcommand{\qtwo}[2][]{%
    \ensuremath{%
        \tilde{\boldsymbol{\theta}}_{#1}^{(#2)}%
    }%
}

\newcommand{\cp}[1]{\left( #1 \right)}
\newcommand{\cb}[1]{\left[ #1 \right]}

\usepackage{lastpage}
\jmlrheading{xx}{2025}{1-\pageref{LastPage}}{x/xx; Revised x/xx}{x/xx}{xx-0000}{Yaomengxi Han and Debarghya Ghoshdastidar}

\ShortHeadings{Attention Learning is Needed to Efficiently Learn Parity Function}{Attention Learning is Needed to Efficiently Learn Parity Function}
\firstpageno{1}

\begin{document}

\title{Transformers Provably Learn Sparse XOR with Polylogarithmic Parameters}

\author{\name Yaomengxi Han \email maxcharm.han@tum.de \\
       \addr School of Computation, Information and Technology\\
       Technical University of Munich\\
       Boltzmannstrasse 3, 85748, Munich, Germany
       \AND
       \name Debarghya Ghoshdastidar \email ghoshdas@cit.tum.de \\
       \addr School of Computation, Information and Technology\\
       Technical University of Munich\\
       Boltzmannstrasse 3, 85748, Munich, Germany}

\editor{My editor}

\maketitle

\begin{abstract}
Learning sparse parity functions has become a theoretical testbed for studying feature learning in neural networks. However, existing analyses primarily focus on Feed-Forward Neural Networks (FFNNs). Meanwhile, theoretical understanding of Transformers in this setting remains limited, despite their empirical success and structural suitability for discovering sparse support over long sequences. To address this gap, we analyze how a single-layer, two-head Transformer learns the sparse XOR problem. Considering samples \((\mathbf{x}, y) \in \{\pm 1\}^d \times \{\pm 1\}\), where the label is defined by \(y = -x_{i^*} x_{j^*}\) for some unknown \(i^*, j^* \in [d]\), we prove that, with only \(\mathcal{O}(\mathrm{polylog}(d))\) trainable parameters, Transformers can successfully discover the relevant features and drive the loss for every input to nearly \(0\) with one gradient step. This result establishes that Transformers break the fundamental \(\Omega(d)\) parameter bottleneck inherent to FFNNs for this problem. Furthermore, we empirically show that this rapid feature discovery is uniquely driven by the exact softmax attention, outperforming common substitutes such as linear or component-wise attention. Finally, we provide a theoretical sample complexity bound for learning from finite data, demonstrating the generalization ability of Transformers in this task.
\end{abstract}


\section{Introduction}
Sparse parity learning has emerged as a standard theoretical tool for understanding feature learning in neural networks~\citep{BarakEGKMZ22, AbbeAM23, sanguino2024}. Consider binary input sequences \(\mathbf x\in \{\pm1\}^d\), these functions output the parity of an unknown target support \(S\) of size \(k \ll d\), formally defined as \(y = -\prod_{i\in S}x_i\). A core challenge for different models in learning such functions lies in \emph{parameter efficiency}, more precisely, how many parameters a model class requires to discover this hidden subset and achieve small errors, given samples \((\mathbf x, y)\) from a distribution agnostic to the target support.

This problem effectively separates the parameter efficiency of two-layer Feed-Forward Neural Networks (FFNNs) from linear classifiers on top of fixed, data-independent embeddings. Specifically, previous work~\citep{DanielyM20} proves that under a slightly biased data distribution, gradient descent allows an FFNN to achieve a small expected hinge loss using \(\Omega(dk^7\log k)\) parameters. In contrast, a linear classifier yields an expected hinge loss close to \(1/2\) unless its embedding dimension scales exponentially with \(k\). From this example, one can clearly see how sparse parity learning cleanly separates models capable of feature learning via a trainable first layer, such as FFNNs, from those trapped in the lazy training regime with static embeddings, such as Neural Tangent Kernels (NTKs).

To make these analyses closer to the ideal and unbiased setting, i.e., \(\mathbf x\sim \mathrm{Unif}(\{\pm 1\}^d)\), extensive prior work has focused on the fundamental sparse XOR problem, where \(k=2\) and \(y = -x_{i^*}x_{j^*}\) for unknown indices \(i^*, j^*\in[d]\). As the simplest parity function, where each individual feature still has no marginal correlation with the label, sparse XOR serves as the canonical theoretical testbed for evaluating the feature learning capabilities of neural networks. In this setting, it has been established that a two-layer FFNN trained with (stochastic) gradient descent can provably achieve the Bayes risk with \(\mathcal{O}(d)\) parameters with a fixed second layer in a noisy setting~\citep{FreiCB23}, or converge with a vanishing risk with \(\mathcal{O}(d\cdot \mathrm{polylog}(d))\) parameters with a trainable output layer in a noiseless setting~\citep{Glasgow24}. These results demonstrate the ability of FFNNs to actively learn sparse information and overcome the representational limits of fixed-feature models.

While FFNNs require fewer parameters than linear classifiers in learning such sparse functions, the required number of parameters still grows at least linearly with the input length \(d\)~\citep{Ghorbani19, DanielyM20, SanfordHT23, Glasgow24}. This linear dependency is a fundamental consequence of their dense, fully connected first layer, which inherently couples the weight dimension to the input size, trapping the architecture in a strict \(\Omega(d)\) parameter bottleneck.

In contrast, Transformers~\citep{Vaswani21} are structure-wise a strong candidate to break this parameter bottleneck, as their weight matrices for attention score calculation are shared across the entire input sequence. By mapping each input bit into an \(m\)-dimensional embedding space, the core parameter matrices (\(W_Q, W_K, W_V \in \mathbb{R}^{m \times m}\)) scale solely with the embedding dimension \(m\), remaining completely independent of the sequence length \(d\). Empirically, Transformers outperform FFNNs in discovering sparse features from high-dimensional ambient inputs~\citep{SanfordHT23, BhattamishraPKB23}. Yet theoretical explanations for this success remain limited, largely due to the difficulty of analyzing the exact dynamics of softmax activation. Consequently, previous work on Transformer learning dynamics often substitutes softmax with simplified variants, such as linear~\citep{AhnCSYJS24, Lu24}, or component-wise~\citep{MarionBBB25} attention.

However, recent work~\citep{DengSYZ25, Duranthon26} shows that the softmax attention used in practice is theoretically superior over other simplified substitutes, both in terms of expressivity and statistical learnability. In terms of expressivity, \citet{DengSYZ25} establishes a strict representational separation. Using a carefully constructed binary classification task, they show that a four-layer softmax network can perfectly distinguish between two synthetic data classes, whereas linear attention falls short. Regarding learnability, \citet{Duranthon26} analyzes a single-location regression task, proving that while softmax attention can achieve a vanishing population error in the high-dimensional limit, other variants, including component-wise attention, fail to do the same. Note that while the single-location regression task only requires the discovery of an individual token, the sparse XOR task is more complex, since it requires modeling the interaction between two relevant bits. To address this, Transformers must not only isolate the target support but also ensure that different attention heads specialize in different positions rather than collapsing onto the same feature. Prior work also investigates solving parity with Transformers~\citep{KimS25} with a focus on Chain-of-Thought (CoT) reasoning. Their analysis relies on one-hot positional encoding to enforce exact pairwise orthogonality between distinct positions. Although this encoding scheme simplifies the analysis, it also forces the embedding dimension to scale linearly with \(d\), which results in \(\mathcal{O}(d^2)\) trainable parameters and offers no advantage over FFNNs regarding parameter efficiency.

Motivated by this, we rigorously analyze the learning dynamics of a single-layer, two-head transformer with softmax attention on the sparse XOR problem. Our main result demonstrates that transformers completely break the \(\Omega(d)\) parameter bottleneck required by FFNNs:
\begin{theorem}[\textbf{Main Result, Informal Version of Theorem~\ref{thm:main}}]\label{theorem:main_result}
For input sequences of length \(d\), a single-layer, two-head Transformer with a fixed output function can \textbf{exactly learn} the sparse XOR function in a single population gradient step, given a suitable learning rate. Specifically, with high probability over the random initialization, for every input sequence \(\mathbf{x} \in \{\pm 1\}^d\), the instance-wise loss converges to \(0\) as \(d\) increases. Critically, this is achieved with an embedding dimension of only \(m = \mathcal{O}(\mathrm{polylog}(d))\), therefore the overall parameter complexity remains \(\mathcal{O}(\mathrm{polylog}(d))\).
\end{theorem}

Our main result yields four key insights. First, it directly implies that Transformers strictly surpass FFNNs in learning sparse XOR in terms of parameter efficiency, reducing the required parameter complexity from \(\Omega(d)\) to \(\mathcal{O}(\mathrm{polylog}(d))\) (Section~\ref{subsec:param_bottleneck}). Second, we empirically compare the attention evolution and loss trajectories of true softmax, linear, and component-wise attention variants, showing that the learning dynamics of exact softmax attention are necessary for this rapid, one-step convergence (Section~\ref{subsec:softmax_necessity}). Third, our analysis reveals that Transformers offer an interpretable approach for sparse feature learning. Specifically, we prove a strict one-to-one alignment between attention heads and the target support during the initial gradient step. This theoretical alignment is supported empirically by the attention evolution through finite-sample gradient descent, demonstrating that Transformers provide useful information about the sparse features in early stage of training (Section~\ref{subsec:attention_interpretability}). Finally, beyond population gradient descent, we establish a theoretical sample complexity bound that proves the Transformer's ability to generalize from finite training data (Section~\ref{subsec:finite-sample}).

\section{Preliminaries}
We formally define the theoretical model used to solve the sparse XOR problem in Section~\ref{subsec:transformer}, and detail the training process, along with the initialization scheme in Section~\ref{subsec:training}. Finally, we establish that our model is mathematically equivalent to a standard Transformer model in Section~\ref{subsec:equivalence}.

\subsection{Model Architecture}\label{subsec:transformer}


\begin{figure}[t]
    \centering
    \includegraphics[width=\linewidth]{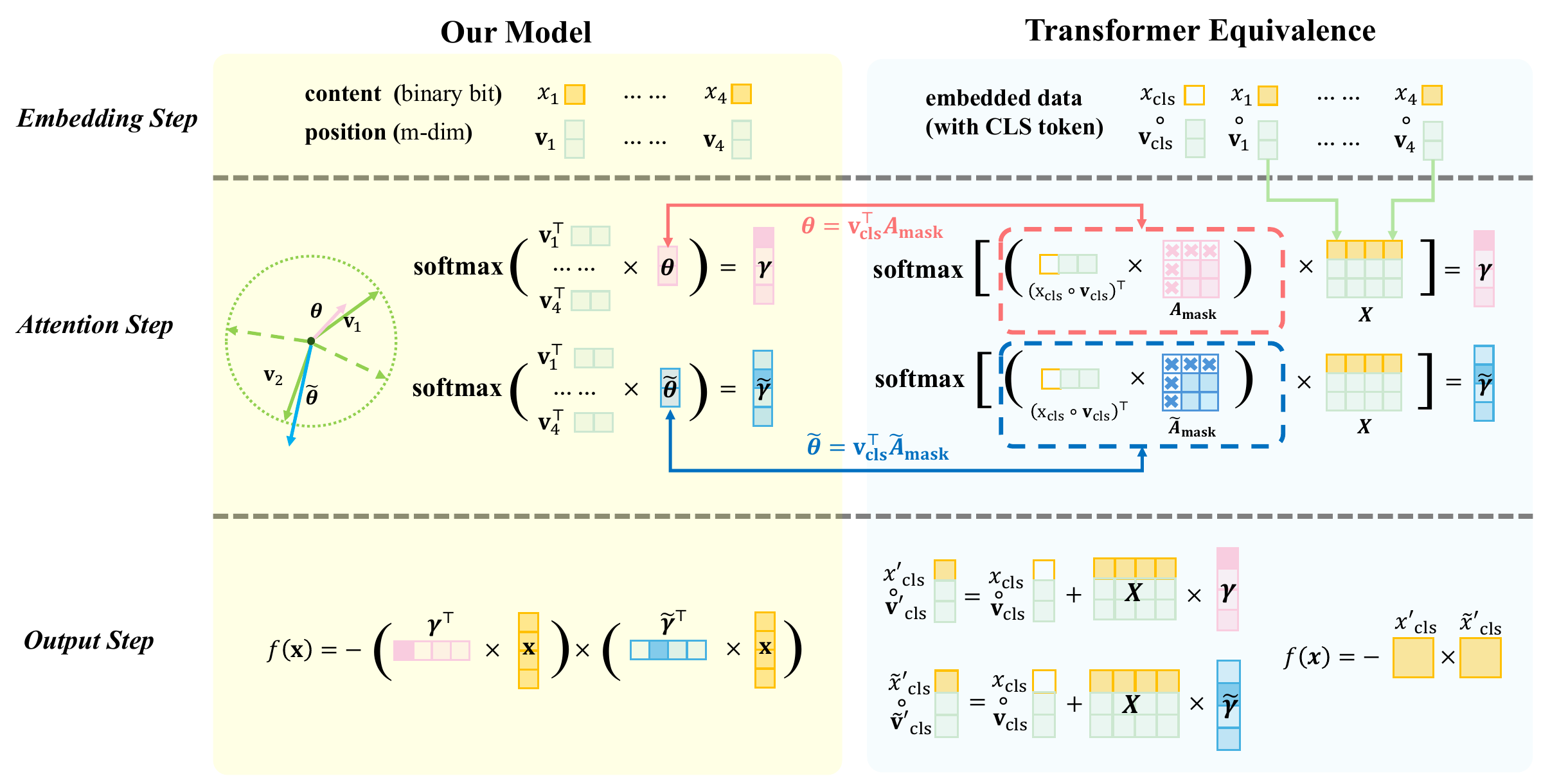}
    \caption{\textbf{Architectural equivalence between our model (left) and a standard single-layer Transformer (right).} Assuming optimized weights, we visualize the forward pass of both models on the input \([x_1, \dots, x_4]^\top\) (illustrated with embedding dimension \(m=2\) and \(S=\{1,2\}\)). In the \textbf{Embedding Step}, our model decouples bit contents \(\{x_i\}_{i=1}^{4}\) from positional embeddings \(\{\mathbf{v}_i\}_{i=1}^4\). The standard Transformer concatenates the content and position vectors and prepends a CLS token to the sequence. In the \textbf{Attention Step}, our model discovers the support target (denoted by the darker locations in the softmax scores \(\boldsymbol{\gamma}\) and \(\boldsymbol{\tilde{\gamma}}\)) via maximizing the alignment between the learnable query vectors \(\boldsymbol{\theta}\), \(\tilde{\boldsymbol{\theta}}\) and the relevant positions \(\mathbf v_1, \mathbf v_2\) (visualized in the green circle). In the standard Transformer, this dynamic is mathematically recovered via the products of the CLS embedding (\(\mathbf v_{\text{cls}}\)) and the attention matrices (\(\mathbf{A_{\text{mask}}}\), \(\tilde{\mathbf{A}}_{\text{mask}}\)), where structural masking (crosses denote \(-\infty\)) forces the CLS token to ignore the content channel and prevents self-attention. In the \textbf{Output Step}, both models compute the prediction by taking the negative product of the content values across heads.}
    \label{fig:model_equiv}
\end{figure}

Figure~\ref{fig:model_equiv} (left column) illustrates the overall architecture of our Transformer model with one attention layer and two attention heads. Recall that the target function for the sparse XOR problem is \(y(\mathbf x) = -x_{i^*}x_{j^*}\) for some unknown support \(i^*, j^*\in[d]\). A natural strategy to solve this problem is to first discover the relevant support and then compute the negative product of the input bits within this support. Consequently, the overall output of our model is formulated as:
\begin{equation} \label{eq:output}
    f(\mathbf x) = -\Biggl(\sum_{i\in[d]} \gamma_i x_i\Biggr) \cdot \Biggl(\sum_{i\in[d]} \tilde\gamma_i x_i\Biggr),
\end{equation}
where \(\gamma_i\) and \(\tilde\gamma_i\) denote the attention weights assigned to position \(i\) by the first and second attention heads respectively. Note that quantities associated with the first head are denoted without a tilde (e.g., \(\gamma_i\)), while those associated with the second are marked with a tilde (e.g., \(\tilde{\gamma}_i\)). The attention weights for both heads are defined respectively as:
\begin{equation}\label{eq:attention}
    \gamma_{i} = \frac{\exp(\mathbf v_i^\top \boldsymbol\theta)}{\sum_{j\in[d]}\exp(\mathbf v_j^\top \boldsymbol{\theta})}, \quad \tilde{\gamma}_i = \frac{\exp(\mathbf v_i^\top \tilde{\boldsymbol\theta})}{\sum_{j\in[d]}\exp(\mathbf v_j^\top \tilde{\boldsymbol{\theta}})}, \quad \forall i\in[d];
\end{equation}
where \(\{\mathbf v_i\}_{i=1}^d\) are the \(m\)-dimensional fixed positional embeddings for each \(i\in[d]\), and \(\boldsymbol{\theta}, \tilde{\boldsymbol{\theta}} \in \mathbb R^m\) are the two query vectors for the first and the second attention head, respectively. Note that the embeddings \(\{\mathbf v_i\}_{i=1}^d\) are \emph{frozen} once initialized. Therefore, the two query vectors are the only parameters updated during training. We detail the architecture via the following steps:

\textbf{Embedding Step.} To discover the target support, our model has to make use of the position information. Therefore, we explicitly inject knowledge regarding the \emph{absolute position} of each bit by assigning a fixed \(m\)-dimensional embedding \(\mathbf v_i\) to each index \(i\in[d]\). This allows the attention heads to locate the relevant features by geometrically aligning with the target embeddings \(\mathbf v_{i^*}\) and \(\mathbf v_{j^*}\). To prevent interference between positions, all embeddings must be approximately pairwise orthogonal. To achieve this efficiently without using high-dimensional one-hot encoding (\(m=d\)), we independently sample \(\mathbf v_i\sim\mathrm{Unif}(\mathbb S^{m-1})\). By scaling \(m\) logarithmically with \(d\), this spherical sampling guarantees approximate orthogonality while maintaining a compact representation (\(m \ll d\)). We provide empirical analysis on how the convergence of our model is affected by the relation of \(m\) and \(d\) in Section~\ref{subsec:param_bottleneck}, and rigorously analyze how the choice of \(m\) affects the residual correlation between these positional embeddings and the final pointwise loss for every input in Section~\ref{sec:proof}.

\textbf{Attention Step.} As formalized in Equation~\ref{eq:attention}, the two learnable query vectors \(\boldsymbol{\theta}, \tilde{\boldsymbol{\theta}} \in \mathbb R^m\) determine the attention score for each position based on how closely they align with the fixed positional embeddings. 
After training, it would be desirable for each query vector to attend to a distinct relevant position in \(S\) with a dominant attention score. This implies a strict one-to-one mapping between the two query vectors and the target positions \(\mathbf{v}_{i^*}\) and \(\mathbf{v}_{j^*}\). Specifically, each head specializes to a distinct relevant position, with its corresponding maximum weight approaching \(1\).

\textbf{Output Step.} The output layer takes the negation of the product of the attention-weighted bit sums from both heads. If the attention score distribution with each head is optimal, then the two sums in Equation~\ref{eq:output} will collapse closely to \(x_{i^*}\) and \(x_{j^*}\). This allows the network output to approximate the true sparse XOR target, yielding \(f(\mathbf{x}) \approx -x_{i^*} x_{j^*}\).

\subsection{Training and Initialization}\label{subsec:training}
We optimize the model using the Mean Squared Error (MSE).\footnote{Although sparse XOR is a classification task, using MSE loss is considered standard in theoretical settings~\citep{KimS25}} For any input sequence \(\mathbf x\in \{\pm 1\}^d\), the corresponding loss regarding the model output is:
\begin{equation}\label{eq:instance_loss}
    \ell(\mathbf{x}) = (y - f(\mathbf{x}))^2 = \cp{-x_{i^*} x_{j^*}+\cp{\sum_{i\in[d]} \gamma_i x_i}\cdot \cp{\sum_{i\in[d]} \tilde\gamma_i x_i}}^2.
\end{equation} 

Let \(\mathcal{D}\) denote the joint data distribution over \((\mathbf{x}, y)\), where the input sequences are sampled uniformly from the \(d\)-dimensional hypercube, \(\mathbf{x} \sim \mathrm{Unif}(\{\pm 1\}^d)\), and the labels are generated by the parity function. We can then formulate the population loss regarding this distribution at step \(t\) as \(\mathcal{L}^{(t)} = \mathbb{E}_{(\mathbf{x}, y) \sim \mathcal{D}} [\ell(\mathbf{x})]\). Let the learning rate be \(\eta > 0\),
then \(\boldsymbol{\theta}\) and \(\tilde{\boldsymbol{\theta}}\) are updated with the following rule:
\begin{equation}\label{eq:param_update}
    \qone{t+1} = \qone{t} - \eta\nabla_{\qone{t}}\mathcal L^{(t)}, \quad \qtwo{t+1} = \qtwo{t} - \eta\nabla_{\qtwo{t}}\mathcal L^{(t)};
\end{equation}
where the bracketed superscript represents the training time step \(t\).

\paragraph{Initialization Assumptions.} Recall that we assume the positional embeddings are sampled uniformly from the unit sphere and \emph{frozen}, i.e., \(\mathbf{v}_i \sim \mathrm{Unif}(\mathbb{S}^{m-1})\). For the trainable parameters, we initialize the query vectors as \(\boldsymbol{\theta}^{(0)} \sim \mathrm{Unif}(\mathbb{S}^{m-1})\) and \(\tilde{\boldsymbol{\theta}}^{(0)} = -\boldsymbol{\theta}^{(0)}\). This ensures the initial softmax scores are approximately uniform across all positions, and provides maximal initial separation to help with head specialization during training. Based on these initialization assumptions, we make two simplifications for the remainder of our analysis: (i) Because the sampling of \(\mathbf{v}_i\) is independent and rotational invariant, we assume the target support is \(S = \{1, 2\}\), hence \(y(\mathbf x)=-x_1x_2\) for any input \(\mathbf x\). (ii) We assume the first query vector is initialized closer to the first target, satisfying \(\mathbf{v}_1^\top \qone{0} > \mathbf{v}_2^\top \qone{0}\). The analysis for the reverse case is mathematically identical and hence omitted.

\subsection{Equivalence to Standard Transformer Architecture}\label{subsec:equivalence}
Our model corresponds to a single-layer, two-head Transformer where the attention layer is leveraged for support discovery (Figure~\ref{fig:model_equiv}, right column). We explain this equivalence in the following steps:

\paragraph{Embedding Equivalence} For an input sequence \(\mathbf{x} = [x_1, \dots, x_d]^\top\), traditional Transformer  concatenates it with a classification (CLS) token \(x_{\text{cls}}\) and \(m\)-dimensional positional embeddings \(\mathbf{v}_i\), yielding the augmented input matrix \(
    \mathbf{X} = \cb{x_{\text{cls}} \circ \mathbf{v}_{\text{cls}}^\top |x_1 \circ \mathbf{v}_1^\top | \dots | x_d \circ \mathbf{v}_d^\top } \in \mathbb{R}^{(1+d) \times (1+m)}\).
To adapt for the sparse XOR problem, we make two explicit design choices that depart from this standard approach. First, because the target support \(S\) is independent of the specific bit contents, we decouple the positional embedding from the bit information, using solely the positional embeddings \(\{\mathbf{v}_i\}_{i=1}^d\) to compute the attention weights (acting as \textbf{Keys}), and the bit content \(x_i \in \{\pm 1\}\) exclusively for the final output (acting as \textbf{Values}). Second, while standard Transformers use sinusoidal embeddings or similar variants to ensure greater correlation between proximate positions, the same proximity bias is not desired for sparse XOR learning. In contrast, we uniformly sample the embeddings from the unit sphere to minimize correlation between different positions.

\paragraph{Attention Equivalence} In standard Transformers, the two attention heads are parameterized by two learnable joint key-query matrices \(\mathbf{A}, \tilde{\mathbf A} \in \mathbb{R}^{(m+1) \times (m+1)}\) (\(\mathbf A =\mathbf W_Q \mathbf W_K^\top\), \(\tilde{\mathbf A} = \tilde{\mathbf W}_Q \tilde{\mathbf W}_K^\top\)), a common technique in attention analysis~\citep{HuangCL24, ChenL25, KimS25}. Since the output layer only considers the updated CLS embeddings from both heads, we define a structural mask on \(\mathbf A\) and \(\tilde{\mathbf A}\) to enforce two constraints: (i) ignoring the content channel, and (ii) preventing the CLS token from attending to itself. The resulting attention weights from both heads for the \(i\)-th position are respectively \(\gamma_i = \frac{\exp(\mathbf{v}_{\text{cls}}^\top \mathbf{A}_{\text{mask}} \mathbf{v}_i)}{\sum_{j=1}^d \exp(\mathbf{v}_{\text{cls}}^\top \mathbf{A}_{\text{mask}} \mathbf{v}_j)}\), and \(\tilde\gamma_i = \frac{\exp(\mathbf{v}_{\text{cls}}^\top \tilde{\mathbf{A}}_{\text{mask}} \mathbf{v}_i)}{\sum_{j=1}^d \exp(\mathbf{v}_{\text{cls}}^\top \tilde{\mathbf{A}}_{\text{mask}} \mathbf{v}_j)}\).
Here, the projected terms \(\mathbf{v}_{\text{cls}}^\top \mathbf{A}_{\text{mask}}\) and \(\mathbf{v}_{\text{cls}}^\top \tilde{\mathbf{A}}_{\text{mask}}\) correspond to the \textbf{queries}. Critically, if we consider updating the matrices \(\mathbf A, \tilde{\mathbf A}\) with gradient descent, then their gradients \(\nabla_{\mathbf{A}} L\) and \(\nabla_{\tilde{\mathbf{A}}} L\) are always projected onto an \(m\)-dimensional manifold because the positional embedding of the CLS token (\(\mathbf{v}_{\text{cls}}\)) remains static. This ensures the learning dynamics of both matrices are mathematically equivalent to the evolution of the reparameterized query vectors \(\boldsymbol{\theta}^\top = \mathbf{v}_{\text{cls}}^\top \mathbf{A}_{\text{mask}}\) and \(\tilde{\boldsymbol{\theta}}^\top = \mathbf{v}_{\text{cls}}^\top \tilde{\mathbf{A}}_{\text{mask}}\). Hence, the matrix-based optimization of the standard Transformer reduces to the vector-based dynamics of our model defined in Equation~\ref{eq:param_update}.

\paragraph{Output Equivalence} To generate the prediction, we restrict the value matrix \(\mathbf{W}_V\) to select only the first dimension of the augmented input \(\mathbf{X}\). This successfully isolates the bit information \(x_i\) for each position \(i\in[d]\). Consequently, the output of the standard attention mechanism recovers the attention-weighted sums \(\sum_{i=1}^d \gamma_i x_i\) and \(\sum_{i=1}^d \tilde\gamma_i x_i\) formalized in Equation~\ref{eq:output}.

\section{Main Result and Discussions}\label{sec:discussion}
Considering the Transformer model described in Section~\ref{subsec:transformer} and the corresponding training scheme in Section~\ref{subsec:training}, we restate our main result formally with the following theorem:

\begin{theorem}[\textbf{Transformers Exactly Learns Sparse XOR in a Single Gradient Step}]
\label{thm:main}
There exist constants\footnote{In Sections~\ref{sec:discussion} and \ref{sec:proof}, letters \(c, C\), and their subscripted or subscripted variants denote positive constants independent of \(d\) and \(m\). Note that their exact values may change across different lemmas and theorems.} \(C, C', C'' > 0\) such that for any positional-embedding dimension \(m \geq (\log d)^4\) and learning rate \(\eta \geq m^{3/2}d^2 \), with probability at least \(1 - m^{-C}\) over the joint random initialization of positional embeddings and query vectors, the Transformer discovers the relevant support with one gradient descent update. Consequently, the instance loss is bounded for any input \(\mathbf{x} \in \{\pm 1\}^d\) by:
\[
    \ell(\mathbf{x}) \le C' d^{-C''}.
\]
\end{theorem}
Theorem~\ref{thm:main} suggests that the instance loss vanishes as \(d\) increases. While we defer the proof to Section~\ref{sec:proof}, we first highlight several theoretical and practical insights of this result. 


\begin{figure}[h]
    \centering
    \begin{subfigure}{0.44\textwidth}
        \centering
        \includegraphics[width=\linewidth]{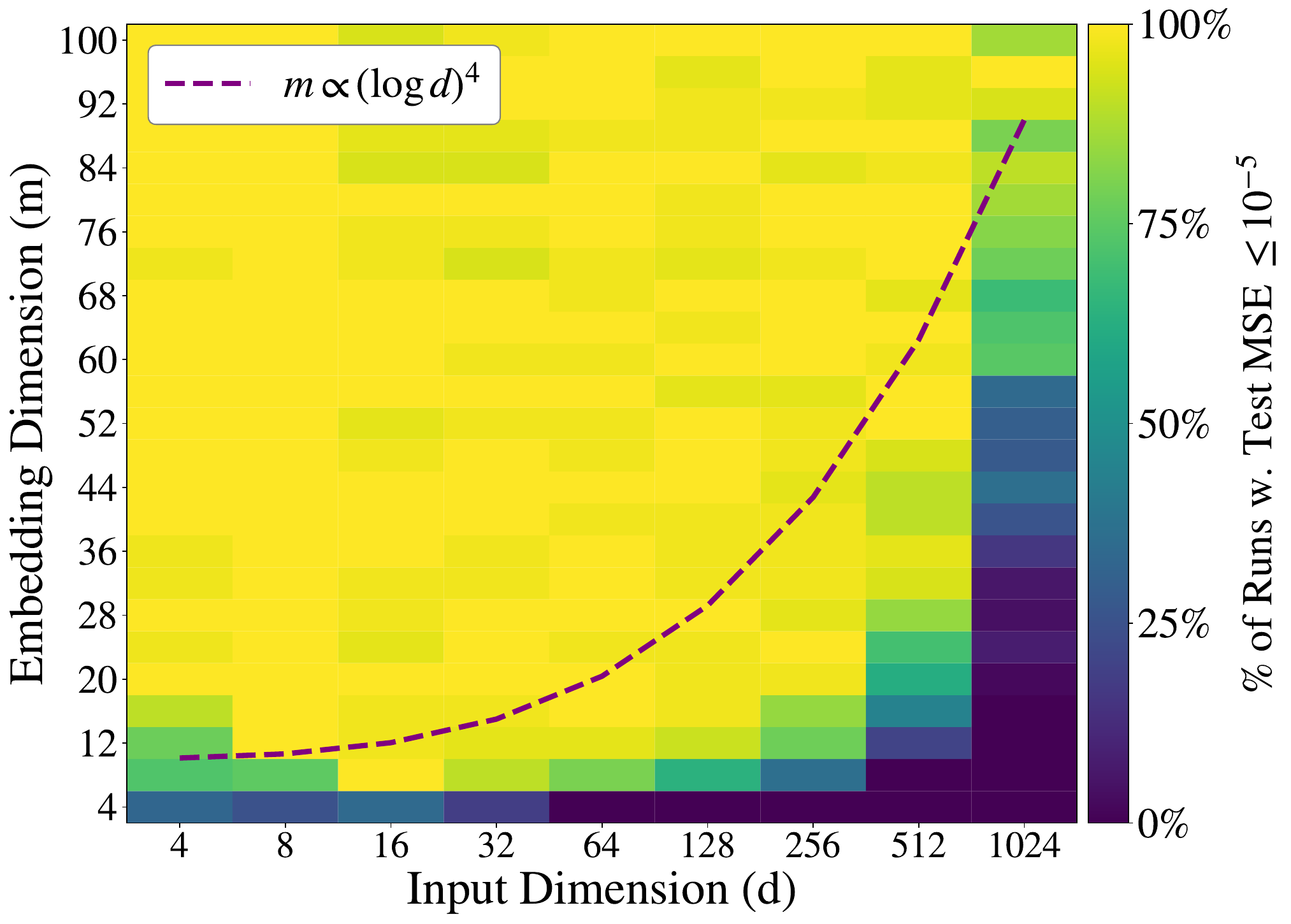}
        \caption{Percentage of Runs with MSE \(\leq 10^{-5}\).}
        \label{fig:heatmap_mse}
    \end{subfigure}
    \begin{subfigure}{0.44\textwidth}
        \centering
        \includegraphics[width=\linewidth]{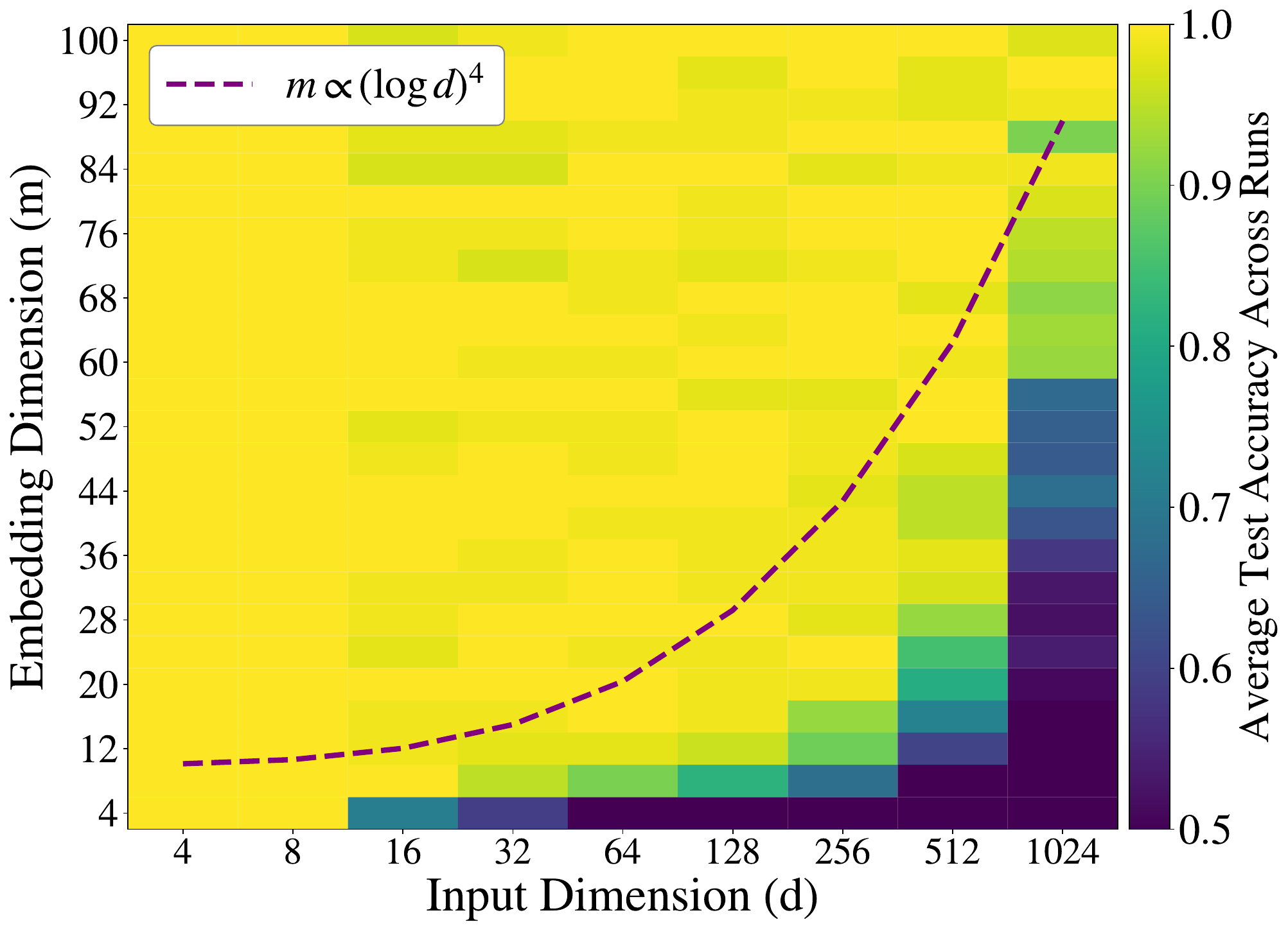}
        \caption{Average Test Accuracy Across All Runs.}
        \label{fig:heatmap_01}
    \end{subfigure}
    \caption{\textbf{Empirical Model Performance Across Different Input and Embedding Lengths.} We evaluate the Transformer over a grid of configurations, scaling the input dimension \(d\) exponentially from \(4\) to \(512\) and the embedding dimension \(m\) from \(4\) to \(100\) with a step size of \(4\). For each \((d, m)\), the model is trained for 20 epochs using full-batch gradient descent on 40,000 samples before evaluated on 20,000 holdout test samples. The heatmaps display the results across 50 runs for each config, displaying the percentage of runs achieving MSE below \(10^{-5}\) (\textbf{left}), and the average accuracy across runs (\textbf{right}). The purple dashed lines in both heatmaps correspond to the theoretical scaling of \(m\), suggested by Theorem~\ref{thm:main}.
    }
    \label{fig:exp_dm}
\end{figure}

\begin{figure}[h]
    \centering
    \begin{subfigure}{0.55\textwidth}
        \centering
        \includegraphics[width=\linewidth]{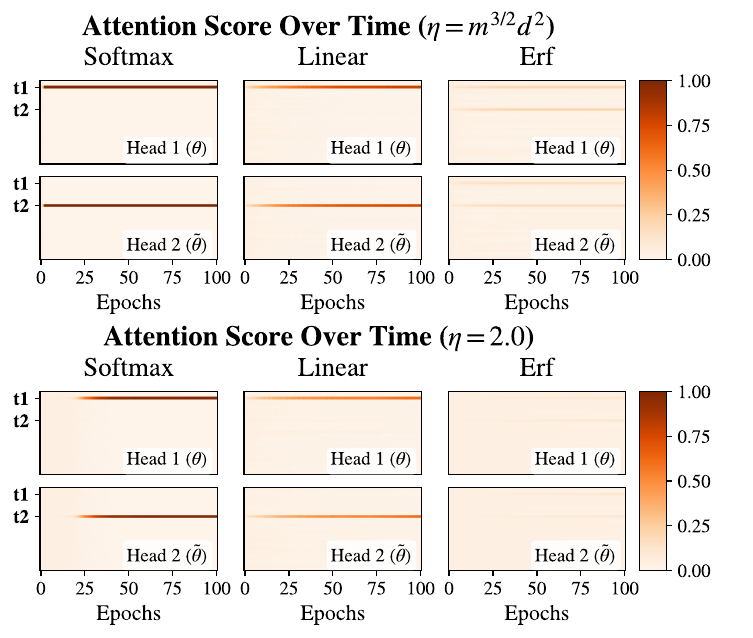}
        \caption{Attention Score Evolution for Two Heads (\(\boldsymbol{\theta}\), \(\tilde{\boldsymbol{\theta}}\))}
        \label{fig:sub_a}
    \end{subfigure}
    \begin{subfigure}{0.41\textwidth}
        \centering
        \includegraphics[width=\linewidth]{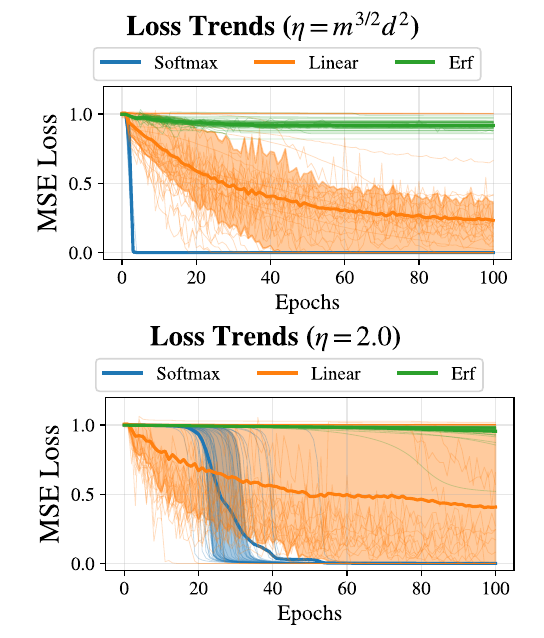}
        \caption{MSE Loss (\(20\)-\(80\) percentiles in shade)}
        \label{fig:sub_b}
    \end{subfigure}
    \caption{\textbf{Feature Learning and Convergence of Transformers with Three Attention Variants.} We evaluate the attention evolution and training loss of Transformers with softmax, linear, and error-function (erf) attention on the sparse XOR task. Across all configurations, we set the input dimension \(d=30\) and embedding dimension \(m=20\). To ensure consistent visualization, target positions are fixed: \(\mathbf {t_1}=2\) and \(\mathbf {t_2}=10\). We report results over \(50\) independent runs, each trained for \(100\) epochs. The \textbf{top row} adopts our theory-driven learning rate (\(\eta = m^{3/2}d^2\)), while the \textbf{bottom row} utilizes a lower, empirical learning rate (\(\eta = 2\)). The \textbf{left column} displays attention evolution heatmaps for both heads, where darker colors indicate higher attention weights on respective positions. The \textbf{right column} plots the corresponding training loss trajectories, denoting the mean (bold line) and the 20th to 80th percentile range (shaded region) across all runs.
    }
    \label{fig:exp_main}
\end{figure}

\subsection{Transformers Break the Parameter Bottleneck of FFNNs}\label{subsec:param_bottleneck}
Prior works~\citep{DanielyM20, Glasgow24} prove that FFNNs lean sparse XOR when the network size scales at least linearly with the input dimension \(d\). 
However, they cannot improve this efficiency because this linear dependency is a fundamental architectural bottleneck in dense networks. 
In contrast, Transformers completely decouple the position embedding dimension \(m\) from the sequence length \(d\), reducing the parameter requirement from \(\Omega(d)\) to \(\mathcal O(\mathrm{polylog}(d))\). To complement this theoretical implication, we empirically investigate the relations among the sequence length \(d\), the embedding dimension \(m\), and the resulting performance of the Transformer. Figure~\ref{fig:exp_dm} presents the overview of the test-time MSE losses and accuracy across various \(d,m\) configurations. Specifically, Subfigure~\ref{fig:heatmap_mse} depicts the percentage of runs (out of 50 independent trials) that successfully converge to a small MSE loss (\(\leq 10^{-5}\)), while Subfigure~\ref{fig:heatmap_01} reports the average accuracy across all runs. As observed in both heatmaps, the Transformer exhibits a clear phase transition: with an embedding dimension of roughly \(m \propto (\log d)^4\), it converges and generalizes reliably, achieving perfect accuracy. This empirical finding reinforces that the number of parameters required for Transformers to solve sparse XOR is exponentially smaller than that for FFNNs.

\subsection{Softmax is Necessary for Efficient Feature Discovery}\label{subsec:softmax_necessity} We show empirically that the one-step convergence in Theorem~\ref{thm:main} relies on the unique dynamics of softmax. Figure~\ref{fig:exp_main} compares our model against the linear and error-function (erf) attention variants. Under the theory-driven learning rate (Figure~\ref{fig:exp_main}, top row), both heads with softmax rapidly align with the relevant support. The heads specialize on distinct target positions (\(\mathbf {t_1}\) and \(\mathbf{t_2}\)), driving the attention scores on these respective positions to nearly \(1.0\) (left column) and achieving loss convergence within the first few epochs in almost all runs (right column). In comparison, the other variants only faintly discover the target bits. In particular, the erf attention fails to specialize across different targets, with a training loss stuck around $0.9$ for most runs. Meanwhile, the losses of linear attention also fluctuate around \(0.5\) and only converge over a much longer horizon in approximately half of the runs. With a lower learning rate commonly used in practice (Figure~\ref{fig:exp_main}, bottom row), softmax-attention heads gradually specialize on different bits, with scores close to \(1.0\). While in some runs, the other variants eventually discover the support, they still fail to converge in more than half of the runs. In conclusion, softmax attention is necessary for both rapid, strong feature alignment and reliable loss convergence.

\subsection{Transformers Offer Interpretable Learning via Head-Position Alignment}\label{subsec:attention_interpretability} During the training process of FFNNs on sparse XOR, the feature discovery relies on a small subset of neurons eventually aligning with the directions of four possible target directions, i.e., \(\{(\pm1,\pm1,0,\dots,0)\}\), assuming \(S = \{1,2\}\)~\citep{FreiCB23, Glasgow24}. However, there is no reliable way to identify which specific neurons are learning the target support before full convergence, for these relevant neurons are obscured by a large population of other irrelevant neurons. Hence, visualizing the network weights to quickly extract information about the target support is difficult. In contrast, Transformers offer information regarding the target support in the first step (see Figure~\ref{fig:sub_a}, top), where the attention distribution of two heads cleanly separates the target positions from the irrelevant positions. The reason behind this clean separation is two-fold: \textbf{(i) Attention Dominance:} For both heads, the gradient alignment with the relevant positions \(k\in\{1,2\}\) is scale-wise larger than with any ambient position \(k'\in[d]\setminus \{1,2\}\). Therefore, for both heads, the sum of the attention scores on two target positions will be close to 1, which we prove rigorously in Lemma~\ref{lm1:scale}. \textbf{(ii) Attention Specialization:} The anti-symmetric initialization for \(\boldsymbol{\theta}\) and \(\tilde{\boldsymbol{\theta}}\) ensures that the two heads specialize to distinct relevant features rather than collapsing onto the same bit (see Lemma~\ref{lm2:symmetry_breaking}). These two phenomena ensure a one-to-one alignment between the query vectors and the target positions. Because the softmax scores collapse onto distinct, relevant positions across different heads after the first step, the evolving attention map provides an intuitive visualization of the feature-learning process. Empirically, even with a smaller learning rate and finite-sample gradient descent, the attention evolution becomes informative within the first few epochs (see Figure~\ref{fig:sub_a}, bottom).

\subsection{Learning with Finite Samples}\label{subsec:finite-sample}
The main focus of this work is on how learning softmax attention in transformers helps to learn sparse XOR below the $\Omega(d)$ parameter bottleneck. To this end, considering the population loss \(\mathcal{L} = \mathbb{E}_{(\mathbf{x}, y) \sim \mathcal{D}} [\ell(\mathbf{x})]\) makes the analysis tractable since the loss simplifies under expectation. However, a complete theoretical picture of sparse feature learning requires extending these results to the practical finite-sample regime. 
To address this, we consider the setting of $n$ training samples $\{(\mathbf{x_i}, y_i)\}_{i\in[n]} \sim_{i.i.d} \mathcal{D}$ and minimization of the empirical loss \(\widehat{\mathcal{L}} = \displaystyle \frac1n \sum\limits_{i=1}^n \ell(\mathbf{x_i})\). The following corollary to Theorem \ref{thm:main} provides a sufficient condition on $n$ required for a single step of full-batch gradient descent to exactly learn Sparse XOR.

\begin{corollary}[\textbf{Sample Size for Learning Sparse XOR with Transformer, complete proof in Appendix~\ref{proof:cor:sample-GD}}]
\label{cor:sample-GD}
There exist absolute constants \(C, C', C'' > 0\), such that for any embedding dimension \(m \geq \log^4 d\), learning rate \(\eta \geq m^{3/2} d^2\), and sample size \(n \geq md^4\), with probability at least \(1 - m^{-C}\) over the training samples and the joint random initialization of the positional embeddings and the query vectors, the Transformer discovers the relevant support after a single gradient descent update. Consequently, the instance loss is bounded for any input \(\mathbf{x} \in \{\pm 1\}^d\) by $\ell(\mathbf{x}) \le C' d^{-C''}$.
\end{corollary}

Corollary \ref{cor:sample-GD} implies that $n = \Omega(d^4\cdot \text{polylog}(d))$ is sufficient to learn sparse XOR using a transformer. We observe that this sample size is worse than the known sample size requirement of $\Omega(d^2)$ for single-step gradient descent on FFNNs \citep{BarakEGKMZ22} or $\Omega(d\cdot \text{polylog}(d))$ with stochastic gradient descent on FFNNs \citep{Glasgow24}. We believe that the sub-optimal sample size requirement from Corollary \ref{cor:sample-GD} is a consequence of our proof, which simply controls the deviation of the gradients of population loss and empirical loss at initialization. 
However, as detailed in Appendix~\ref{sec: sample_complexity_empirical}, our empirical results show that Transformers generalize with significantly fewer samples in practice. We leave the question of improving sample complexity for Transformers as an open problem. 

\section{Proof of Main Theorem}\label{sec:proof}
     We outline the key steps in proving Theorem~\ref{thm:main} and defer the full proofs of all supporting lemmas to Appendices~\ref{proof:lm1} and \ref{proof:lm2}. As discussed in Section~\ref{subsec:attention_interpretability}, two critical phenomena occur simultaneously during the initial gradient step. First, both heads attend exclusively to the target support, driving the attention scores on all ambient positions to nearly $0$. Second, the query vectors of the two heads learn to align with distinct relevant positions. We begin by formalizing the first phenomenon.
    
    \begin{lemma}[\textbf{Gradient Separation between Relevant-Ambient Positions, complete proof in Appendix~\ref{proof:lm1}}]
    \label{lm1:scale}
        Let the embedding dimension \(m \in \mathrm{polylog}(d)\). Considering the gradient for both query vectors, there is a scale-wise separation between their alignment with relevant and ambient positions. Formally, there exist constants \(c_1,c_1',c_1''>0\), such that with probability at least \(1-d^{-c_1}\), we have:
        \begin{align*}
            & \min_{k\in \{1,2\}} -\langle \nabla_{\qone{0}} \mathcal L^{(0)},  \mathbf v_k \rangle \geq \frac{2d^{-2}}{\exp(4)} - \frac{c_1'}{d^2}\sqrt{\frac{\log d}{m}}; \quad \max_{k'\in[d]\setminus \{1,2\}} -\langle \nabla_{\qone{0}} \mathcal L^{(0)},  \mathbf v_{k'} \rangle \leq \frac{c_1''}{d^2}\sqrt{\frac{\log d}{m}}; \\
            & \min_{k\in \{1,2\}} -\langle \nabla_{\qtwo{0}} \mathcal L^{(0)},  \mathbf v_k \rangle \geq \frac{2d^{-2}}{\exp(4)} - \frac{c_1'}{d^2}\sqrt{\frac{\log d}{m}}; \quad \max_{k'\in[d]\setminus \{1,2\}} -\langle \nabla_{\qtwo{0}} \mathcal L^{(0)},  \mathbf v_{k'} \rangle \leq \frac{c_1''}{d^2}\sqrt{\frac{\log d}{m}}.
        \end{align*}
    
    \end{lemma}
    From Lemma~\ref{lm1:scale} implies that for both heads, the dominant term in alignment between the negative gradient and the relevant position embeddings is of order \(\Theta \cp{d^{-2}}\), while the maximum alignment between the negative gradient and any irrelevant position is upper bounded by \(\mathcal O\cp{d^{-2} \sqrt{\frac{\log d}{m}}}\). Recall from Equation~\ref{eq:param_update} that this difference is amplified by the scale of the learning rate, and further reflected in the attention scores. Therefore, we have the following result, which states that with a suitable learning rate, the sum of the attention scores on the relevant bits for both heads is close to 1. 
    
    \begin{corollary}[\textbf{Attention Dominance on Relevant Positions, complete proof in Appendix~\ref{proof:coro_2}}]\label{corollary_dominance}
        Let the embedding dimension be \(m\geq \cp{\log d}^4\) and the learning rate be \(\eta \geq m^{3/2}d^2\), there exist absolute constants \(C_1, C_1'\), such that after the first gradient step, for any \(C_1''>0\), it follows:
    \begin{equation}\label{eq:prob_1}
        \text{with probability at least } 1- d^{-C_1}, \; \atone[1]{1} + \atone[2]{1} \geq 1-C_1' d^{-C_1''}, \; \attwo[1]{1} + \attwo[2]{1} \geq 1-C_1' d^{-C_1''}.
    \end{equation}
    \end{corollary}

    We can see that as \(d\) increases, \(C_1' \cdot d^{-C_1''}\) shrinks rapidly for large \(C_1''\), therefore the attention weights will concentrate entirely on the target support. Following this, we show that the initialization asymmetry will be amplified in the first gradient step, and will further drive a strict one-to-one alignment between query heads and distinct relevant positions:

    \begin{lemma}[\textbf{Amplification of Initial Asymmetry, complete proof in Appendix~\ref{proof:lm2}}] 
    \label{lm2:symmetry_breaking}
        After one gradient step, each query vector's initial alignment bias between the relevant bits is amplified. Formally, there exist constants \(c_2, c_2', c_2''>0\) such that with probability at least \(1-m^{-c_2}\), we have: 
        \[
            \min \left\{ \frac{\atone[1]{1}}{\atone[2]{1}}, \frac{\attwo[2]{1}}{\attwo[1]{1}} \right\} \geq \exp\cp{\eta \cp{\frac{c_2'}{d^2m} - c_2''\frac{\sqrt{\log d}}{d^2m^{3/2}}}}.
        \]
    \end{lemma}

    Specifically, with a proper learning rate, Lemma~\ref{lm2:symmetry_breaking} directly implies the following:
    \begin{corollary}[\textbf{One-to-One Alignment between Query Vectors and Relevant Positions}]
        Consider \(m\geq \cp{\log d}^4\) and \(\eta \geq m^{3/2}d^2\), there exist constants \(C_2, C_2'>0\), such that after the first gradient descent step, for any \(C_2''>0\), we have:
        \begin{equation}\label{eq:prob_2}
            \text{with probability greater than }1-m^{-C_2}, \;  \frac{\atone[1]{1}}{\atone[2]{1}} \geq C_2'd^{C_2''} \text{ and } \frac{\attwo[2]{1}}{\attwo[1]{1}} \geq C_2'd^{C_2''}.
        \end{equation}
    \end{corollary}
    \begin{proof}
        Plugging \(m = \cp{\log d}^4\) and \(\eta = m^{3/2}d^2\) into Lemma~\ref{lm2:symmetry_breaking} yields the conclusion.
    \end{proof}

    Combining statements \ref{eq:prob_1} and \ref{eq:prob_2} and using union bound over their respective probability, the following result holds: there exists absolute constant \(C, C''>0\), such that with probability at least \(1-m^{-C}\), \(\atone[1]{1} \geq 1-d^{-C''}\) and \(\attwo[2]{1} \geq 1- d^{-C''}\). We shorthand \(\sum_{i\ne 1} \atone[i]{1} x_i\) as \(R\) and \(\sum_{i\ne 2} \attwo[i]{1} x_i\) as \(\tilde R\), and it holds for any input \(\mathbf x\) that:
    \begin{equation*}
        |R| = \left|\sum_{i\ne 1} \atone[i]{1} x_i\right| \leq d^{-C''}, \quad |\tilde R| = \left|\sum_{i\ne 2} \attwo[i]{1} x_i\right| \leq d^{-C''}.
    \end{equation*}
     Plugging these inequalities into Equation~\ref{eq:instance_loss} allows us to bound the loss for any input \(\mathbf x\) by:
      {\small
    \begin{align*}
        \ell(\mathbf x) & = (x_1x_2)^2 + \cp{\sum_{i\in[d]} \atone[i]{1} x_i}^2\cp{\sum_{i\in[d]} \attwo[i]{1} x_i}^2 - 2x_1x_2 \cp{\sum_{i\in[d]} \atone[i]{1} x_i}\cdot \cp{\sum_{i\in[d]} \attwo[i]{1} x_i} \\
        & = 1 + \cp{\atone[1]{1}x_1 + R}^2\cp{\attwo[2]{1}x_2 + \tilde R}^2 - 2x_1x_2\cp{\atone[1]{1}x_1 + R}\cp{\attwo[2]{1}x_2 + \tilde R}\\
        & \overset{(a)}{=} 1 + \cp{\atone[1]{1}\attwo[2]{1}x_1x_2}^2 - 2\atone[1]{1}\attwo[2]{1}x_1^2 x_2^2 + \mathcal{O}\big(|R| + |\tilde R|\big) \\
        & = \cp{1 - \atone[1]{1}\attwo[2]{1}}^2 + \mathcal{O}\big(|R| + |\tilde R|\big) \leq C' d^{-C''},
    \end{align*}
    }
\noindent where equality (a) follows from absorbing the cross terms into the asymptotic notation, based on \(|\atone[1]{1}x_1| \leq 1\) and \(|\attwo[2]{1}x_2| \leq 1\). This completes the proof of Theorem~\ref{thm:main}.

\section{Conclusion}\label{sec:conclusion}
In this work, we prove that Transformers can learn the sparse XOR problem with \(\mathcal O(\mathrm{polylog}(d))\) parameters, surpassing the \(\Omega(d)\) requirement by FFNNs. Beyond this parameter efficiency, we provide insights into Transformers' ability for sparse feature learning, including empirical evidence for the necessity of softmax for rapid convergence, and finite-sample bounds to guarantee generalization. Further discussion comparing the related literature and our contribution is provided in Appendix~\ref{sec:related_works}.


\textbf{Limitations and Future Work.} While we establish the parameter efficiency of Transformers, we acknowledge the following limitations in our analysis. First, the finite-sample complexity bound derived in Section~\ref{subsec:finite-sample} is likely loose. Based on the analytical results in Appendix~\ref{sec: sample_complexity_empirical}, we hypothesize that the sample complexity could be significantly improved, potentially by a more refined analysis on the lower bound of the correlation between positional embeddings. Second, while our asymmetric initialization guarantees head specialization for the sparse XOR problem, it cannot be trivially extended to general \(k\)-parity. For \(k > 2\), Transformers may require more heads to ensure initial asymmetry. Therefore, the fixed output layer used in our analysis is no longer applicable. We provide preliminary experiments regarding this generalized setting in Appendix~\ref{sec:general_k_parity}.

\newpage

\bibliography{arxiv_version}

\newpage

\appendix

\section{Complementary Empirical Results}\label{sec:empirical_results}
\subsection{Finite Sample Regime: Generalization Performance vs. Sample Complexity}\label{sec: sample_complexity_empirical}

\begin{figure}[h]
    \centering
    \begin{subfigure}{0.46\textwidth}
        \centering
        \includegraphics[width=\linewidth]{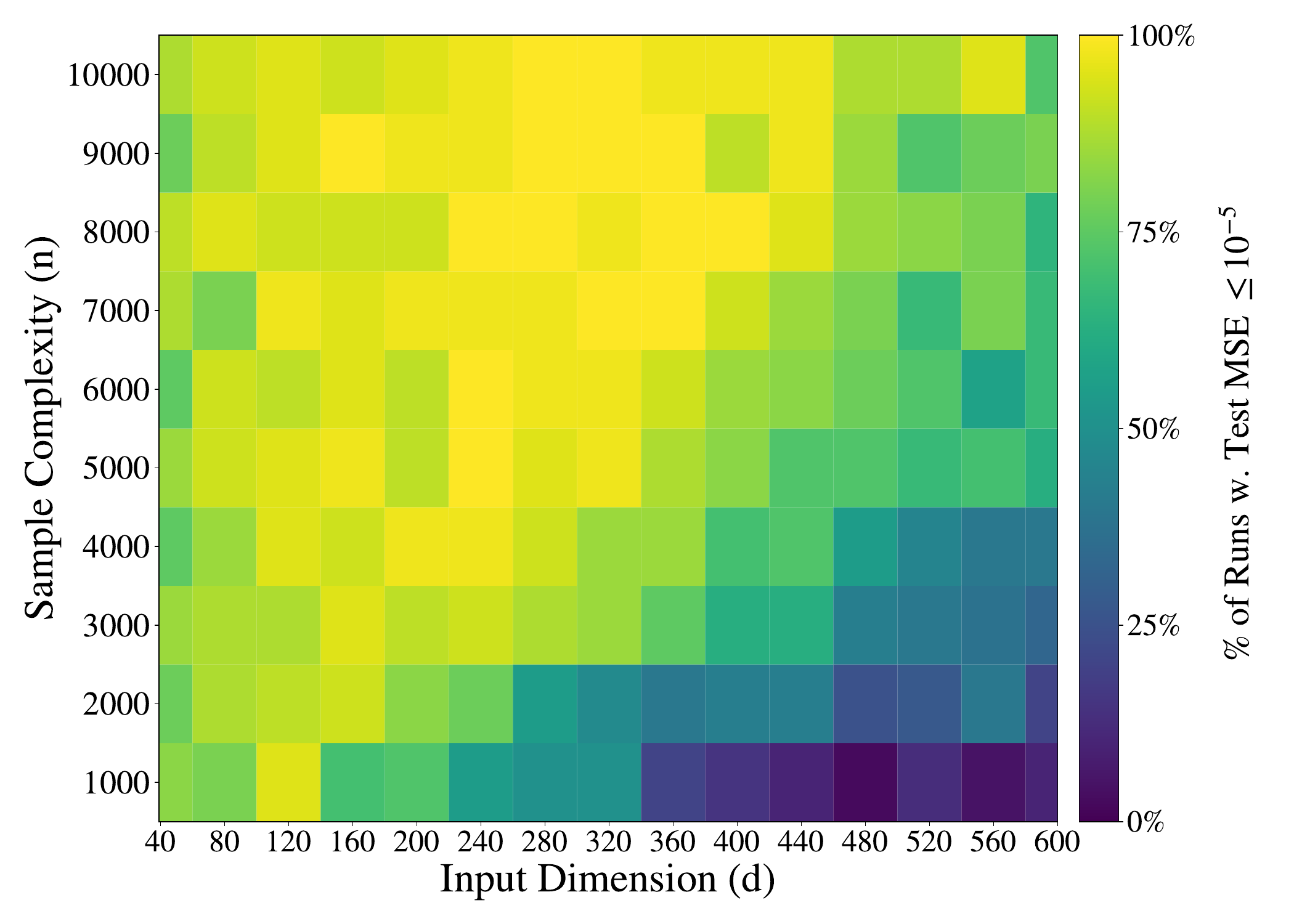}
        \caption{Percentage of Runs with MSE \(\leq 10^{-5}\).}
        \label{fig:sample_mse}
    \end{subfigure}
    \begin{subfigure}{0.46\textwidth}
        \centering
        \includegraphics[width=\linewidth]{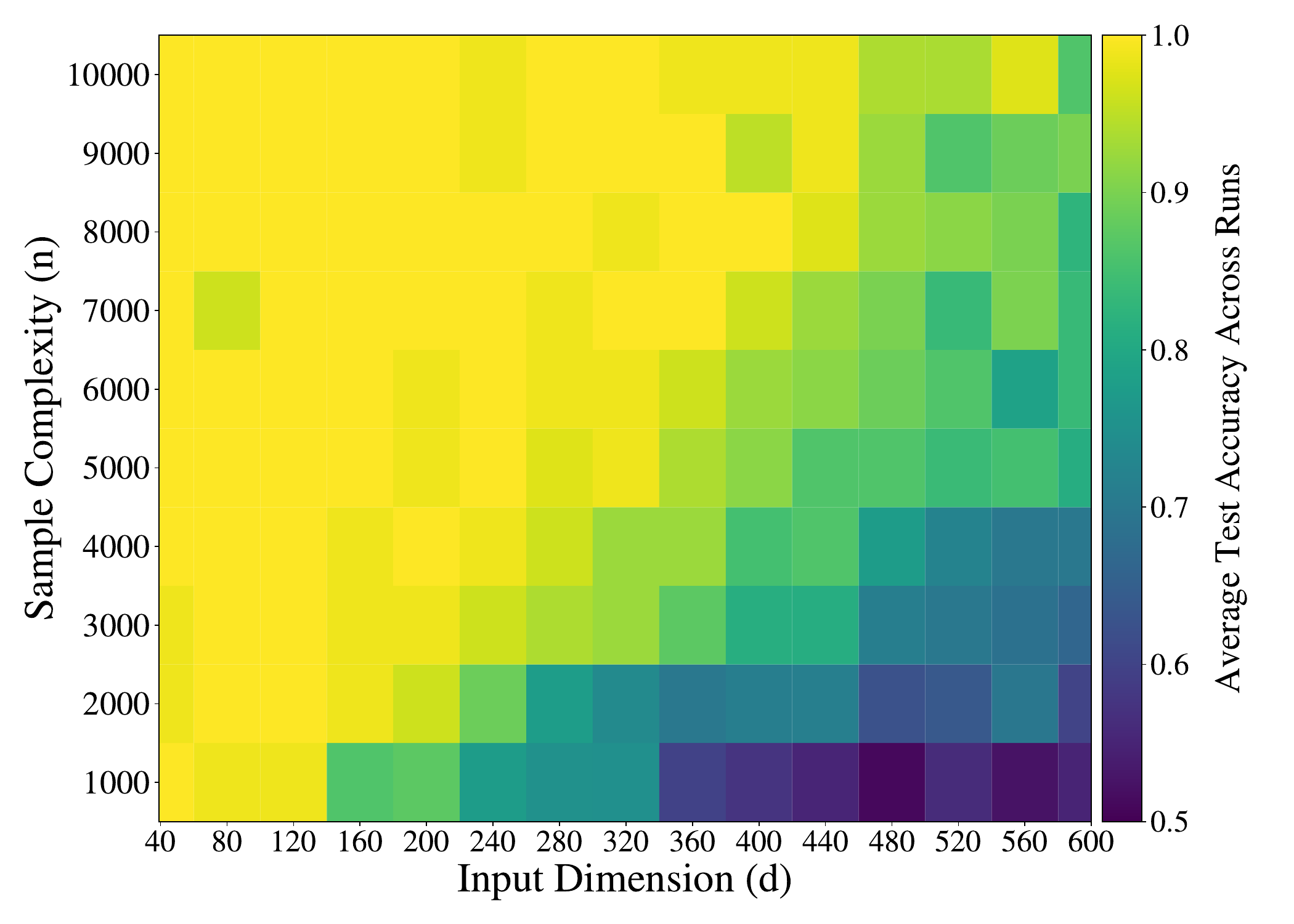}
        \caption{Average Test Accuracy Across All Runs.}
        \label{fig:sample_acc}
    \end{subfigure}
    \caption{\textbf{Empirical Performance Across Different Input Dimension and Sample Complexity.} To empirically validate the required sample complexity for Transformer in learning sparse XOR, we evaluate the model by scaling the input dimension \(d\) from \(40\) to \(600\) linearly, and the number of training samples \(n\) also linearly from \(1,000\) to \(10,000\) with a step size of \(1,000\). The embedding dimension is fixed at \(m=60\) for each configuration. For each \((d, n)\) pair, the model is trained for 20 epochs using full-batch GD for 40 steps and evaluated on 20,000 test samples. The heatmaps display the results across 50 runs for each configuration, displaying respectively the percentage of runs achieving a test MSE below \(10^{-5}\) (\textbf{left}), and the test accuracy across all runs (\textbf{right}). 
    }
    \label{fig:exp_dn}
\end{figure}
Figure~\ref{fig:exp_dn} shows the Transformer model's generalization performance under different input dimension \(d\) and sample complexity \(n\). Both heatmaps exhibit a visible phase transition, with a clear boundary separating the failure (dark green) and success (bright yellow) regimes. It is worth noting that within the regime of our tested configurations, the required number of samples to achieve low MSE and high accuracy scales approximately linearly with the input dimension \(d\). This \(\mathcal{O}(d)\) empirical scaling provides strong evidence that our derived theoretical bounds are conservative and could potentially be tightened in future work.

\subsection{Beyond Sparse XOR: Transformer's Performance on Generalized \(k\)-parity}\label{sec:general_k_parity}
\begin{figure}[h]
    \centering
    \begin{subfigure}{0.48\textwidth}
        \centering
        \includegraphics[width=\linewidth]{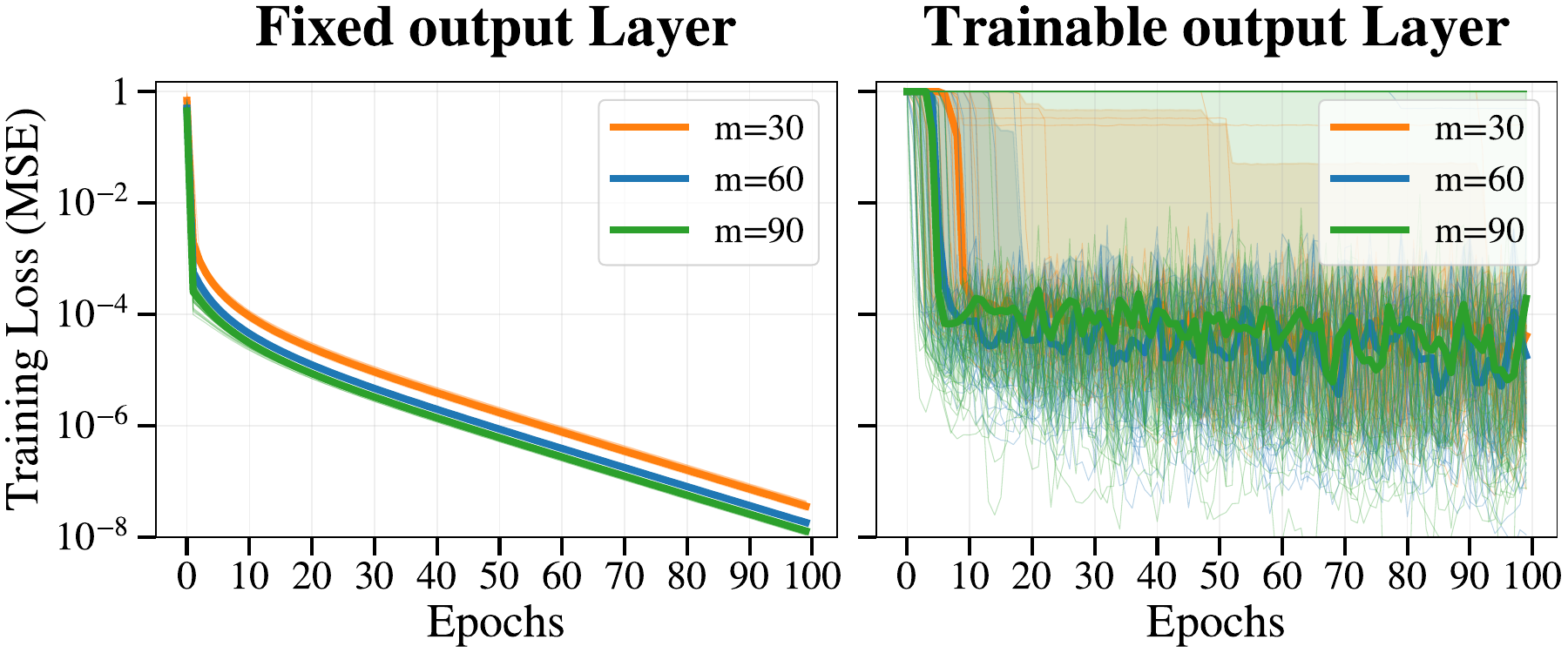}
        \caption{Training Loss of Both Models (\(k=3\), \(d=40\))}
        \label{fig:3_parity_40}
    \end{subfigure}
    \begin{subfigure}{0.48\textwidth}
        \centering
        \includegraphics[width=\linewidth]{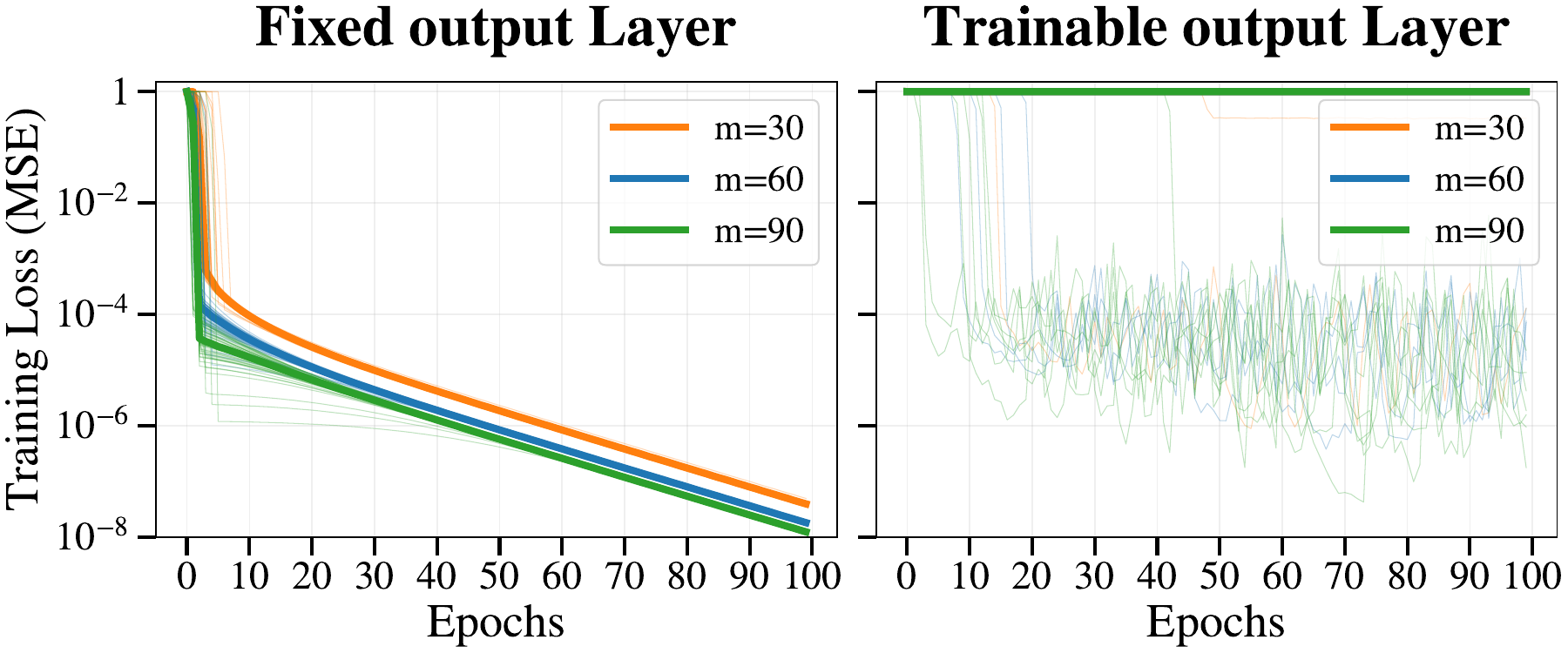}
        \caption{Training Loss of Both Models (\(k=3\), \(d=100\))}
        \label{fig:3_parity_100}
    \end{subfigure}
\caption{\textbf{Comparison of Fixed-output-Layer Transformer versus Trainable-Output Transformer on 3-Parity Learning.} Both Transformer models are trained over 100 epochs for $d=40$ (left) and $d=100$ (right). Each subfigure contrasts the MSE loss trends of a 3-head Transformer using a fixed negative-product as output (left cells) against those of a 6-head Transformer with a trainable two-layer FFNN (right cells) across embedding dimensions \(m \in \{30, 60, 90\}\). Results are aggregated over 50 independent runs, displaying all individual loss trajectories, the median (bold lines), and the 20th–80th percentiles (shaded regions).}
\label{fig:3_parity}
\end{figure}
Figure \ref{fig:3_parity} illustrates the training dynamics of a 3-head Transformer with a fixed negative-product as output compared to a Transformer with a trainable output layer on the sparse 3-parity task. Across both evaluated input dimensions (\(d=40\) and \(d=100\)), the fixed-output Transformer converges rapidly, showing its ability to quickly discover the target support even without an asymmetric initialization scheme. In contrast, the trainable-output architecture exhibits high sensitivity to both input and embedding dimensions. While it converges on some runs when \(d=40\) across all embedding dimensions, it struggles to locate the relevant features when \(d=100\), especially when the embedding dimension is small. Convergence occurs only when $m$ is increased to 90.

\section{Helper Lemmas}
\begin{lemma}[\textbf{Concentration Bounds at Initialization}]
\label{helper_lemma:1}
Denote \(\delta_{ij} = \mathbf v_i^\top \mathbf v_j\), for any embedding dimension \(m\in\mathrm{polylog}(d)\) the following bounds hold:
\begin{enumerate}
    \item For any constant \(c>0\), with probability greater than \(1-d^{-c}\), we have:
    \[
        |\delta_{ij}| \leq  \sqrt{\frac{c\log d}{2m}}, \quad i\ne j;
    \]
    \item There exists constants \(c', c'' > 0\), such that with probability greater than \(1-d^{-c'}\), we have:
    \[
        \left| \sum_{j \ne i} \atone[j]{0} \delta_{ij} \right| \leq  \frac{c''\sqrt{\log d}}{m^{3/2}}.
    \]
    \item There exists constants \(c', c'' > 0\), such that with probability greater than \(1-d^{-c'}\), we have:
    \[
        \left| \sum_{j \ne i} \attwo[j]{0} \delta_{ij} \right| \leq  \frac{c''\sqrt{\log d}}{m^{3/2}}.
    \]
\end{enumerate}
\end{lemma}
\begin{proof}
    The inner product of two spherical vectors is sub-gaussian with \(\sigma_{ij}^2 = \mathrm{var}\cb{\mathbf v^\top_i \mathbf v_j} = \mathbb E\cb{\mathrm{trace}\cp{\mathbf v_i\mathbf v_i^\top\mathbf v_j\mathbf v_j^\top}} = \frac{1}{m}\) and \(\mathbb E\cb{\mathbf v_i^\top \mathbf v_j} = 0\). Therefore, for any fixed constant \(c>0\):
    \[
        \mathbb P\cp{|\delta_{ij}-0|\geq t}\leq \exp\cp{\frac{-2t^2}{\sigma^2}} \implies \mathbb P\cp{|\delta_{ij}|\leq \sqrt{\frac{c\log d}{2m}}}\geq 1-d^{-c}.
    \]
    This concludes the proof for the first bound.
    
    \vspace{3pt}
    
    For the second bound, we first shorthand the following terms:
    \[ 
        z_{-i} = \sum_{j\ne i} \exp(\mathbf v^\top_j \qone{0}), \quad \mathbf w_{-i} = \sum_{j\ne i} \exp(\mathbf v^\top_j \qone{0}) \mathbf v_j.
    \]
    Then the LHS of the second inequality can be bounded deterministically by:
    \begin{equation}\label{eq:h1_b1}
        \left|\sum_{j\ne i}\atone[j]{0} \delta_{ij}\right| = \left|\frac{\mathbf v_i^\top \mathbf w_{-i}}{z_{-i} + \exp(\mathbf v_i^\top \qone{0})}\right| \leq \left|\frac{\mathbf v_i^\top  \mathbf w_{-i}}{z_{-i}}\right|.
    \end{equation}
    W.l.o.g. we can assume that \(\qone{0}\) is fixed. Since \(\mathbf v_i\) is independent on \(z_{-i}\) and \(\mathbf w_{-i}\), we have that with any constant \(c>0\), with probability at least \(1-d^{-c}\):
    \begin{equation}\label{eq:h1_b2}
        \left|\mathbf v_i^\top \mathbf w_{-i}\right| \leq \left\|\mathbf w_{-i}\right\| \cdot  \sqrt{\frac{c\log d}{2m}},
    \end{equation}
    since \(\left|\mathbf v_i^\top \mathbf w_{-i}\right|\) is the inner product of the spherical vector \(\mathbf v_i\) and an independent vector \(\mathbf w_{-i}\), which is sub-gaussian with \(\sigma^2 = \|\mathbf w_{-i}\|^2 / m\). We are going to bound the norm of \(\mathbf w_{-i}\) next. Decomposing \(\mathbf v_j = (\mathbf v_j^\top \qone{0})\qone{0} + \mathbf v_{j, \perp}\), we can rewrite \(\exp(\mathbf v_j^\top \qone{0}) \mathbf v_j\) as:
    \begin{align*}
       \mathbb E\cb{\exp(\mathbf v_j^\top \qone{0}) \mathbf v_j } & = \mathbb E\cb{\exp(\mathbf v_j^\top \qone{0}) (\mathbf v_j^\top \qone{0})\qone{0} } + \mathbb E\cb{\exp(\mathbf v_j^\top \qone{0}) \mathbf v_{j, \perp} } \\
       & \overset{(a)}{=} \mathbb E\cb{\exp(\mathbf v_j^\top \qone{0}) (\mathbf v_j^\top \qone{0})\qone{0} } \\
       & = \mathbb E\cb{p_j \exp(p_j)} \qone{0},
    \end{align*}
    where equation \((a)\) is from the symmetry of \(\mathbf v_{j, \perp}\), and \(p_j = \mathbf v_j^\top \qone{0}\) in the last line is the shorthand for the length of the projection onto the first query vector. For any fixed \(\qone{0}\), the density function of a uniformly-sampled spherical vector's projection onto it can be written as:
    \begin{equation}\label{eq:projection_density}
        g(p_j) = C_m (1 - p_j^2)^{\frac{m-3}{2}}, \quad \forall p_j \in [-1, 1]
    \end{equation}
    where \(C_m = \frac{\Gamma(m/2)}{\sqrt{\pi} \Gamma((m-1)/2)}\) is some normalizing constant. Then we have:
    \begin{align*}
        \mathbb E\cb{p_j \exp(p_j)} & = C_m \cdot \int_{-1}^{1} p_j \exp(p_j)(1-p_j^2)^{\frac{m-3}{2}} \mathrm{d} p_j \\
        & = -\frac{C_m}{m-1} \cdot \int_{-1}^{1} e^x\cp{\frac{\mathrm{d}}{\mathrm{d x}}(1-x^2)^{\frac{m-1}{2}}} \mathrm{d}x \\
        & = -\frac{C_m}{m-1} \cdot \cp{\cb{e^x (1-x^2)^{\frac{m-1}{2}}}^{1}_{-1} - \int_{-1}^{1} e^x (1-x^2)^{\frac{m-1}{2}} \mathrm{d}x} \\
        & = \frac{C_m}{m-1} \cdot \int_{-1}^{1} e^x (1-x^2)^{\frac{m-1}{2}} \mathrm{d}x \\
        & \leq \frac{1}{m-1} \cdot \int_{-1}^{1} e^x C_m (1-x^2)^{\frac{m-3}{2}} \mathrm{d}x \\
        & = \frac{1}{m-1}\mathbb E\cb{e^{p_j}} \leq \frac{1}{m-1}\exp\cp{\frac{1}{2(m-1)}} \leq \frac{2}{m-1}.
    \end{align*}
    Due to symmetry argument we can easily see that \(\mu_j = \mathbb E\cb{p_j \exp(p_j)} \geq 0\). Therefore, we have \(\mu_j \in [0, \frac{2}{m-1}]\), and hence the expectation can be bounded by:
    \[
        \left\|\mathbb E\cb{\mathbf w_{-i} }\right\| = \left|\sum_{j\ne i} \mu_j \qone{0}\right| = \sum_{j\ne i} \mu_j \left\|\qone{0}\right\| \leq \frac{2(d-1)}{m-1}.
    \]
    With any fixed \(\qone{0}\), we can view \(\mathbf w_{-i} = \sum_{j\ne i} \exp(p_j) \mathbf v_j\) as the sum of \(d-1\) pairwise independent vectors. We use the shorthand \(G(\mathbf v_{-i}) = \left\|\mathbf w_{-i}  - \mathbb E \cb{\mathbf w_{-i}}\right\|\) and the difference between \(\left|G(\dots, \mathbf v_k, \dots) - G(\dots, \mathbf v_k', \dots)\right|\) for any \(k \in [d] \setminus \{i\}\) can be bounded by:
    \[
        \left|G(\dots, \mathbf v_k, \dots) - G(\dots, \mathbf v_k', \dots)\right| \leq \left\|\exp(\mathbf v_k^\top \qone{0}) \mathbf v_k - \exp(\mathbf v_k'^\top \qone{0}) \mathbf v_k'\right\| \leq 2e,
    \]
    where the first inequality stems from triangle inequality and the second is from the fact that both \(\mathbf v_k\) and \(\mathbf v_k'\) are from the unit sphere. Then by Mcdiarmid inequality we have, for any constant \(c>0\), there exists some constant \(c'>0\) such that:
    \[
        \mathbb P\cp{|G(\mathbf v_{-i}) - \mathbb E\cb{G(\mathbf v_{-i})}| \geq t} \leq \exp\cp{\frac{-2t^2}{\sum_{j\ne i}\cp{2e}^2}} \implies \mathbb P\cp{|G(\mathbf v_{-i}) - \mathbb E\cb{G(\mathbf v_{-i})}| \geq c\sqrt{d \log d}} \leq d^{-c'},
    \]
    In addition, if we denote \(Y_j = \exp(p_j)\mathbf v_j - \mathbb E\cb{\exp(p_j)\mathbf v_j}\), and for any fixed \(\qone{0}\), \(Y_j\) and \(Y_k\) are independent when \(j\ne k\), therefore we have:
    \begin{align*}
        \mathbb E\cb{G(\mathbf v_{-i})} & = \mathbb E\cb{\left\|\sum_{j\ne i} Y_j\right\|} \leq \sqrt{\mathbb E \cb{\left\|\sum_{j\ne i} Y_j\right\|^2}} \\
        & = \sqrt{\mathbb E\cb{\sum_{j\ne i}Y_j^\top Y_j + \sum_{j\ne i}\sum_{k\notin \{i, j\}} Y_j^\top Y_k}} \\
        & = \sqrt{\sum_{j\ne i} \mathbb E\cb{Y_j^\top Y_j} + \sum_{j\ne i}\sum_{k\notin \{i, j\}} \mathbb E[Y_j]^\top \mathbb E[Y_k]} \\
        & = \sqrt{\sum_{j\ne i} \mathbb E[\left\|Y_j\right\|^2]} = \sqrt{\sum_{j\ne i} \mathbb E \cb{\left\|\exp(p_j)\mathbf v_j - \mathbb E\cb{\exp(p_j)\mathbf v_j}\right\|^2}} \\
        & = \sqrt{\sum_{j\ne i} \cp{\mathbb E \cb{\left\|\exp(p_j)\mathbf v_j\right\|^2} - \left\|\mathbb E\cb{\exp(p_j)\mathbf v_j}\right\|^2}} \leq \sqrt {\sum_{j\ne i}  \mathbb E\cb{\exp(2p_j)\left\|\mathbf v_j\right\|^2}} \leq e\cdot \sqrt{(d-1)}
    \end{align*}
    Due to the triangle inequality, we have for any constant \(c >0\), there exists some constants \(c', c''>0\) such that with probability larger than \(1-d^{-c'}\):
    \begin{align}
        \|\mathbf w_{-i}\| & \leq  \left\|\mathbb E\cb{\mathbf w_{-i}}\right\| + \left\|\mathbf w_{-i}  - \mathbb E \cb{\mathbf w_{-i}}\right\| \notag \\
        & \leq \left\|\mathbb E\cb{\mathbf w_{-i} }\right\| + \mathbb E\cb{G(\mathbf v_{-i})} + \left|G(\mathbf v_{-i}) - \mathbb E\cb{G(\mathbf v_{-i})}\right|  \notag \\
        & \leq \frac{2(d-1)}{m-1} + e\sqrt{d-1} + c\sqrt{d\log d}  \notag \\
        & \leq c\sqrt{d\log d} + c'' \frac{d}{m} \leq (c+c') \frac{d}{m}.\label{eq:h1_b3}
    \end{align}
    Lastly we lower bound \(z_{-i}\), the expected value of which is:
    \[
        \mathbb E\cb{z_{-i}} = \sum_{j\ne i}\mathbb E\cb{\exp(\mathbf v_j^\top \qone{0})} = (d-1) \cdot \mathbb{E} \cb{\exp(\mathbf v_j^\top \qone{0})} \overset{(b)}{\geq} (d-1) \cdot \exp\cp{\mathbb E[\mathbf v_j^\top \qone{0}]} = d - 1.
    \]
    The inequality \((b)\) is from Jensen's inequality. Besides, since it is the sum of \(d-1\) independent random variables bounded within \([-e, e]\), using Hoeffding's inequality we have, for any constant \(c_1 > 0\), there exists some \(c_1' > 0\) such that:
    \[
        \mathbb P\cp{|z_{-i} - \mathbb E[z_{-i}]|\geq c_1\sqrt{d\log d}} \leq d^{-c_1'}.
    \]
    Therefore, for any constant \(c_1>0\), there exists \(c_1' > 0\) such that with probability more than \(1-d^{-c_1'}\), we get \(\displaystyle z_{-i} \geq c_1 d\). Combining this with bounds from Equations \ref{eq:h1_b1}, \ref{eq:h1_b2} and \ref{eq:h1_b3} together, and applying the union bound yields the second inequality.

    The proof of the third inequality is the same as the second due to the symmetry for \(\qone{0}\) and \(\qtwo{0}\).
    
\end{proof}

\begin{lemma}[\textbf{Anti-Concentration Bound: Initial Gap between Relevant positions}]\label{helper_lemma:2}
    For any \(c>0\), with probability at least \(1-m^{-c}\) over the joint initialization assumption, we have:
    \[
        \frac{\atone[1]{0}}{\atone[2]{0}} = \frac{\attwo[2]{0}}{\attwo[1]{0}} \geq 1+ m^{-(0.5+c)}.
    \]
\end{lemma}
\begin{proof}
    Consider the definition of the attention scores, we have:
    \[
        \frac{\atone[1]{0}}{\atone[2]{0}} = \exp\cp{\mathbf v_1^\top \qone{0}  - \mathbf v_2^\top \qone{0}} = \exp\cp{\mathbf v_2^\top \cp{-\qone{0}} - \mathbf v_2^\top \cp{-\qone{0}}} = \frac{\attwo[2]{0}}{\attwo[1]{0}}.
    \]
    Hence, the equality holds in the conclusion. Focus on the first query vector, it follows that:
    \begin{equation}\label{eq:h2_final}
        \frac{\atone[1]{0}}{\atone[2]{0}} = \exp\cp{\mathbf v_1^\top \qone{0}  - \mathbf v_2^\top \qone{0}} \geq 1+ \cp{\mathbf v_1^\top \qone{0}  - \mathbf v_2^\top \qone{0}}
    \end{equation}
    Denote \(z=\mathbf v_2^\top\qone{0}\), and use \(g(z)\) to denote the density function of \(z\) as defined in Eq.~\ref{eq:projection_density}. Given that \qone{0} favours \(\mathbf v_1\) over \(\mathbf v_2\), we can upper bound the probability of \(\mathbf v_1^\top\qone{0} - z\) being smaller than \(\delta\) as:
    \begin{equation}\label{eq:helper_2}
       \mathbb P\cp{\mathbf v_1^\top\qone{0} - z< \delta \, | \, \mathbf v_1^\top\qone{0} > z} = 2\int_{-1}^{1} P( \mathbf v_1^\top\qone{0} < z+\delta) g(z) dz \leq 2\delta \sqrt{\frac{m}{2\pi}},
    \end{equation}
    which comes from the fact that the density of \(\mathbf v_1^\top \qone{0}\) is bounded by \(\sqrt{\frac{m}{2\pi}}\) everywhere. Substitute \(\delta = \frac{1}{m^{c+0.5}}\) in Eq.~\ref{eq:helper_2} and rearrange, we get that with any \(c>0\):
    \[
        \mathbb P\cp{\mathbf v_1^\top \qone{0} - \mathbf v_2^\top \qone{0} \geq \frac{1}{m^{0.5+c}}} \geq 1 - \sqrt{\frac{2}{\pi}} m^{-c}.
    \]
    Plugging Eq.~\ref{eq:h2_final} into the inequality above yields the complete conclusion.
\end{proof}
\section{Complete Proof of Lemma~\ref{lm1:scale}}\label{proof:lm1}
\begin{proof}
    At initialization, the population gradient regarding the first query vector \qone{0} is:
    \begin{align*}
        \nabla_{\qone{0}} \mathcal L^{(0)} & = \sum_{i=1}^d \fracpartial{\mathcal L}{\atone[i]{0}}\cdot \nabla_{\qone{0}} \atone[i]{0} \\
        & = -2\cb{\attwo[2]{0}\cdot \atone[1]{0}\sum_{j \ne 1}\atone[j]{t}(\mathbf v_1 - \mathbf v_j) + \attwo[1]{0}\cdot \atone[2]{0}\sum_{j \ne 2}\atone[j]{0}(\mathbf v_2 - \mathbf v_j)} \\
        & \; \quad + 2\|\attwo{0}\|^2_2\cdot\sum_{i=1}^d \cp{\atone[i]{0}}^2 \cdot \sum_{j\ne i} \atone[j]{0}(\mathbf v_i - \mathbf v_j) + 4 \left\langle\atone{0}, \attwo{0}\right\rangle\cdot \sum_{i=1}^d\atone[i]{0}\attwo[i]{0}\sum_{j \ne i} \atone[j]{0}(\mathbf v_i - \mathbf v_j) \\
        & \; \quad - 4 \sum_{i=1}^d \cp{\atone[i]{0}\attwo[i]{0}}^2 \cdot \sum_{j \ne i} \atone[j]{0} (\mathbf v_i - \mathbf v_j), 
    \end{align*}
    Let \(\delta_{ij} = \mathbf v_i^\top \mathbf v_j\). We write the gradient alignment with any position \(k\in[d]\) as the sum of four terms:
    \begin{align*}
        -\left\langle \nabla_{\qone{0}} \mathcal L^{(0)}, \; \mathbf v_k \right\rangle = 
        & \; 2\cb{\attwo[2]{0}\atone[1]{0}\sum_{j \ne 1}\atone[j]{0}(\delta_{1k} - \delta_{jk}) + \attwo[1]{0}\atone[2]{0}\sum_{j \ne 2}\atone[j]{0}(\delta_{2k} - \delta_{jk})}\\
        & \; - 2\|\attwo{0}\|^2_2\cdot\sum_{i=1}^d \cp{\atone[i]{0}}^2 \cdot \sum_{j\ne i} \atone[j]{0}(\delta_{ik} - \delta_{jk})\\
        & \; - 4 \left\langle\atone{0}, \; \attwo{0}\right\rangle\cdot \sum_{i=1}^d\atone[i]{0}\attwo[i]{0}\sum_{j \ne i} \atone[j]{0}(\delta_{ik} - \delta_{jk}) \\
        & \; + 4 \sum_{i=1}^d \cp{\atone[i]{0}\attwo[i]{0}}^2 \cdot \sum_{j \ne i} \atone[j]{0} (\delta_{ik} - \delta_{jk}).
    \end{align*}
    \paragraph{The first term.} Let \(T_1(k) = 2\cb{\attwo[2]{0}\atone[1]{0}\sum_{j \ne 1}\atone[j]{0}(\delta_{1k} - \delta_{jk}) + \attwo[1]{0}\atone[2]{0}\sum_{j \ne 2}\atone[j]{0}(\delta_{2k} - \delta_{jk})}\), we discuss this value in two cases. When \(k=1\), with the conclusion in Helper Lemma~\ref{helper_lemma:1}, there exist absolute constant \(C_1, C_1'\), such that with probability \(1-d^{-C_1}\), we have:
    \begin{align}
        T_1(1) = & \; 2 \attwo[2]{0}\atone[1]{0} \cp{1-\atone[1]{0}} - 2 \attwo[2]{0}\atone[1]{0} \sum_{j\ne 1} \atone[j]{0} \delta_{1j} \notag \\
        & \quad + 2 \attwo[1]{0}\atone[2]{0} \cb{\cp{1-\atone[2]{0}} \delta_{12} - \atone[1]{0} - \sum_{j\notin \{1,2\}} \atone[j]{0} \delta_{1j}} \notag\\
        \geq & \; \frac{2}{d^2\cdot \exp(4)} -\frac{\exp(6)}{d^3} - 2 \attwo[2]{0}\atone[1]{0} \left|\sum_{j \ne 1} \atone[j]{0} \delta_{1j}\right| \notag \\
        & \quad - 2 \attwo[1]{0}\atone[2]{0}|\delta_{12}| -\frac{2\exp(6)}{d^3} - 2\attwo[1]{0}\atone[2]{0} \left|\sum_{j \notin\{1,2\}} \atone[j]{0} \delta_{1j}\right|  \notag \\
        \geq & \; \frac{2}{\exp(4)} \cdot d^{-2} - \frac{C_1'}{d^2}\sqrt{\frac{\log d}{m}} \label{eq:lm1_t1_relevant}.
    \end{align}
    The same applies for \(T_1(2)\). For \(k\notin \{1,2\}\), there exists \(C_1'', C_1'''\), such that w.p. at least \(1-d^{C_1''}\):
    \begin{align}
        T_1(k) = & \; 2 \attwo[2]{0}\atone[1]{0} \cp{1-\atone[1]{0}} \delta_{1k} - 2\attwo[2]{0}\atone[1]{0}\atone[k]{0} - 2\attwo[2]{0}\atone[1]{0}\sum_{j\notin\{1,k\}}\atone[j]{0} \delta_{jk} \notag\\
         & \; + 2\attwo[2]{0}\atone[2]{0} \cp{1-\atone[2]{0}}\delta_{2k} - 2\attwo[1]{0}\atone[2]{0}\atone[k]{0} - 2\attwo[1]{0}\atone[2]{0}\sum_{j\notin\{2,k\}}\atone[j]{0} \delta_{jk} \notag\\
         \leq & \; \frac{2\exp(4)}{d^2}\cdot |\delta_{1k}| + \frac{2\exp(4)}{d^2}\cdot \cp{\left|\sum_{j\notin \{1,k\}}\atone[j]{0}\delta_{jk}\right| + \left|\sum_{j\notin \{2,k\}}\atone[j]{0}\delta_{jk}\right|} + \frac{2\exp(4)}{d^2}\cdot |\delta_{2k}| \notag\\
         \leq & \; \frac{C_1'''}{d^2}\sqrt{\frac{\log d}{m}}. \label{eq:lm1_t1_ambient}
    \end{align}

    \paragraph{The second term.} Let \(T_2(k) = - 2\|\attwo{0}\|^2_2\cdot\sum_{i=1}^d \cp{\atone[i]{0}}^2 \cdot \sum_{j\ne i} \atone[j]{0}(\delta_{ik} - \delta_{jk})\). For any \(k\in[d]\), there exists \(C_2, C_2'>0\), such that with probability at least \(1-d^{-C_2}\), we have:
    \begin{align}
        |T_2(k)| = &\; 2\|\attwo{0}\|^2_2 \cdot \Biggl| \cp{\atone[k]{0}}^2 \cb{\cp{1-\atone[k]{0}} - \sum_{j\ne k}\atone[j]{0} \delta_{jk}} \notag \\
        &\quad + \sum_{i\ne k}\cp{\atone[i]{0}}^2 \cb{\cp{1-\atone[i]{0}}\delta_{ik}-\atone[k]{0}-\sum_{j\notin \{i,k\}}\atone[j]{0}\delta_{jk}} \Biggr| \notag \\
        \leq & \; \frac{\exp(4)}{d} \Biggl[ \frac{\exp(2)}{d^2} \cdot \cp{1+\left|\sum_{j \ne k}\atone[j]{0}\delta_{jk}\right|} + \left|\sum_{i\ne k} \cp{\atone[i]{0}}^2 \delta_{ik}\right| \notag \\
        &\quad + \left|\sum_{i\ne k} \cp{\atone[i]{0}}^3 \delta_{ik}\right| + \frac{\exp(6)}{d^2} + \sum_{i\ne k}\cp{\atone[i]{0}}^2 \left|\sum_{j\notin\{i,k\}} \atone[j]{0} \delta_{jk}\right| \Biggr] \notag \\
        \leq & \; \frac{\exp(4)}{d} \Biggl[ \frac{\exp(2)}{d^2} \cdot \cp{1+\left|\sum_{j \ne k}\atone[j]{0}\delta_{jk}\right|} + \frac{\exp(6)}{d^2} \notag \\
        &\quad + \cp{\max_{j\ne k} \atone[j]{0} + \max_{j\ne k} \cp{\atone[j]{0}}^2} \left|\sum_{i\ne k} \atone[i]{0} \delta_{ik}\right| + \sum_{i\ne k}\cp{\atone[i]{0}}^2 \left|\sum_{j\notin\{i,k\}} \atone[j]{0} \delta_{jk}\right| \Biggr] \notag \\
        \leq & \; C_2' \cdot \frac{\sqrt {\log d}}{d^2 m^{3/2}}. \label{eq:lm1_t2}
    \end{align}
    \paragraph{The third term.} Let \(T_3(k) = - 4 \left\langle\atone{0}, \; \attwo{0}\right\rangle\cdot \sum_{i=1}^d\atone[i]{0}\attwo[i]{0}\sum_{j \ne i} \atone[j]{0}(\delta_{ik} - \delta_{jk})\), for any \(k\in[d]\), there exists \(C_3, C_3'>0\), such that with probability at least \(1-d^{-C_3}\), it follows that:
    
    \begin{align}
        |T_3(k)| = & \; 4\left|\left\langle \atone{0}, \; \attwo{0} \right\rangle\right|\cdot \Biggl|\atone[k]{0}\attwo[k]{0}\cb{\cp{1-\atone[k]{0}}-\sum_{j\ne k}\atone[j]{0}\delta_{jk}} \notag \\
        & \qquad \qquad \qquad \qquad +\sum_{i\ne k}\atone[i]{0}\attwo[i]{0}\cb{\cp{1-\atone[i]{0}}\delta_{ik}-\atone[k]{0}-\sum_{j\notin \{i,k\}}\atone[j]{0}\delta_{jk}}\Biggr| \notag \\
        \leq & \; \frac{4\exp(4)}{d}\cdot \Biggl(\frac{\exp(4)}{d^2}+\frac{\exp(4)}{d^2} \left|\sum_{j\ne k}\atone[j]{0}\delta_{jk}\right| \notag \\
        & \qquad \qquad \qquad +\cp{\max_{i\ne k}\attwo[i]{0}}\cdot \left|\sum_{i\ne k}\atone[i]{0}\delta_{ik}\right| + \frac{\exp(4)}{d}\left|\sum_{i\ne k}\atone[i]{0}\delta_{ik}\right|\Biggr) \notag \\
        \leq & \; C_3' \frac{\sqrt{\log d}}{d^2 m^{3/2}} \label{eq:lm_1_t3}.
    \end{align}
    \paragraph{The fourth term.} Finally, consider \(T_4(k) = 4 \sum_{i=1}^d \cp{\atone[i]{0}\attwo[i]{0}}^2 \cdot \sum_{j \ne i} \atone[j]{0} (\delta_{ik} - \delta_{jk})\), there exists \(C_4, C_4' >0\) such that with probability at least \(1-d^{-C_4}\), we have:
    \begin{align}
        |T_4(k)| = & \; \Biggl|4\cp{\atone[k]{0}\attwo[k]{0}}^2\cb{\cp{1-\atone[k]{0}}-\sum_{j\ne k}\atone[j]{0}\delta_{jk}} \notag \\
        & \quad +4\sum_{i\ne k}\cp{\atone[i]{0}\attwo[i]{0}}^2\cb{\cp{1-\atone[i]{0}}\delta_{ik}-\atone[k]{0}-\sum_{j\notin \{i,k\}}\atone[j]{0}\delta_{jk}} \Biggr| \notag \\
        \leq & \; \frac{4\exp(8)}{d^4} + \frac{4\exp(8)}{d^4}\left|\sum_{j \ne k}\atone[j]{0}\delta_{jk}\right| \notag \\
        & \quad + 4 \cp{\max_{i\ne k} \attwo[i]{0}}^2 \cp{\max_{i\ne k} \atone[i]{0}} \cdot \left|\sum_{i\ne k}\atone[i]{0}\delta_{ik}\right| + 4\sum_{i\ne k}\frac{\exp(8)}{d^4}\left|\sum_{j\notin \{i,k\}}\atone[j]{0}\delta_{jk}\right| \notag \\
        \leq & \; C_4' \frac{\sqrt{\log d}}{d^3 m^{3/2}}. \label{eq:lm1_t4}
    \end{align}
    We take the union bound of Equations \ref{eq:lm1_t1_relevant}, \ref{eq:lm1_t2}, \ref{eq:lm_1_t3} and \ref{eq:lm1_t4} for \(k\in\{1,2\}\), and will get that there exists \(c, c_1'>0\), such that with probability at least \(1-d^{-c}\), we have:
    \begin{equation}\label{eq:lm1_final_relevant}
        \min_{k\in\{1,2\}} -\left\langle\nabla_{\qone{0}}\mathcal L^{(0)}, \mathbf v_k\right\rangle \geq \frac{2d^{-2}}{\exp(4)} - \frac{c_1'}{d^2}\sqrt{\frac{\log d}{m}}.
    \end{equation}
    Again take the union bound of Equations \ref{eq:lm1_t1_ambient}, \ref{eq:lm1_t2}, \ref{eq:lm_1_t3} and \ref{eq:lm1_t4} for any \(k'\in [d]\setminus\{1,2\}\), we have there exists \(c'', c''' > 0\), such that with probability at least \(1-d^{-c''}\), we have:
    \begin{align*}
        -\left\langle\nabla_{\qone{0}}\mathcal L^{(0)}, \mathbf v_{k'}\right\rangle \leq \frac{c'''}{d^2}\sqrt{\frac{\log d}{m}}.
    \end{align*}
    Taking the union bound for all \(k' \in [d]\setminus \{1,2\}\), the \(\log d\) term is absorbed, hence we will have that there exists \(c', c_1'' > 0\) such that with probability at least \(1-d^{-c'}\), we have:
    \begin{equation}\label{eq:lm1_final_ambient}
        \max_{k'\in[d]\setminus \{1,2\}}-\left\langle\nabla_{\qone{0}}\mathcal L^{(0)}, \mathbf v_{k'}\right\rangle \leq \frac{c_1''}{d^2}\sqrt{\frac{\log d}{m}}.
    \end{equation}
    Choose \(c_1\) as smaller constant between \(c\) and \(c'\) from Equation~\ref{eq:lm1_final_relevant} and \ref{eq:lm1_final_ambient} yields the first conclusion in the Lemma. The proof for the second conclusion with \qtwo{0} is identical due to symmetry.
    
\end{proof}
\section{Complete Proof of Corollary~\ref{corollary_dominance}}\label{proof:coro_2}
Recall the definition of softmax score, for the first attention head, we can write the reciprocal of the sum of its attention scores for positions 1 and 2 as:
    \begin{equation}\label{proof_coro2_reciprocal}
        \frac{1}{\atone[1]{1} + \atone[2]{1}} = \frac{\sum_{i\in[d]} \exp\cp{\mathbf v_i^\top \qone{1}}}{\exp\cp{\mathbf v_1^\top \qone{1}} + \exp\cp{\mathbf v_2^\top \qone{1}}} = 1 + \frac{\sum_{i\ne\{1,2\}} \exp\cp{\mathbf v_i^\top \qone{1}}}{\exp\cp{\mathbf v_1^\top \qone{1}} + \exp\cp{\mathbf v_2^\top \qone{1}}}.
    \end{equation}
    We first lower bound \(\exp\cp{\mathbf v_1^\top \qone{1}} + \exp\cp{\mathbf v_2^\top \qone{1}}\) from the definition. By Lemma~\ref{lm1:scale}, we know that there exists \(c_1, c_1'\), such that with probability greater than \(1-d^{-c_1}\), we have:
    \begin{align}
        & \; \exp\cp{\mathbf v_1^\top \qone{1}} + \exp\cp{\mathbf v_2^\top \qone{1}} \notag \\
        = & \;\exp\cp{\mathbf v_1^\top \qone{0}} \cdot \exp\cp{-\eta \left\langle \nabla_{\qone{0}} \mathcal L^{(0)}, \mathbf v_1 \right\rangle} + \exp\cp{\mathbf v_2^\top \qone{0}} \cdot \exp\cp{-\eta \left\langle \nabla_{\qone{0}} \mathcal L^{(0)}, \mathbf v_2 \right\rangle} \notag \\
        \geq & \; \cb{\exp\cp{\mathbf v_1^\top \qone{0}} + \exp\cp{\mathbf v_2^\top \qone{0}}} \cdot \exp\cp{\eta \cp{\frac{2d^{-2}}{\exp(4)} - \frac{c_1'}{d^2}\sqrt{\frac{\log d}{m}}}} \notag \\
        \geq & \; \frac{2}{\exp(2)}\cdot \exp\cp{\eta \cp{\frac{2d^{-2}}{\exp(4)} - \frac{c_1'}{d^2}\sqrt{\frac{\log d}{m}}}}. \label{eq:coro2_relevant}
    \end{align}
    Similarly for \(\sum_{i\ne\{1,2\}} \exp\cp{\mathbf v_i^\top \qone{1}}\), there exists constant \(c_1''>0\), with probability at least \(1-d^{-c_1}\),  we have:
    \begin{align}
        \sum_{i\ne\{1,2\}} \exp\cp{\mathbf v_i^\top \qone{1}} & \leq (d-2) \max_{i\ne\{1,2\}} \exp\cp{\mathbf v_i^\top \qone{1}} \notag \\
        & \leq (d-2)\cdot \exp(2) \cdot \exp\cp{\eta\cp{\frac{c_1''}{d^2}\sqrt{\frac{\log d}{m}}}} \label{eq:coro2_ambient}
    \end{align}
    Taking the ratio of Equation \ref{eq:coro2_ambient} and \ref{eq:coro2_relevant}, plug in \(\eta \geq m^{3/2}d^2\) and \(m\geq \cp{\log d}^4\), such that for large enough \(d\), there exists some constant \(c>0\) such that for any \(c'>0\):
    \begin{align*}
        \frac{\sum_{i\ne\{1,2\}} \exp\cp{\mathbf v_i^\top \qone{1}}}{\exp\cp{\mathbf v_1^\top \qone{1}} + \exp\cp{\mathbf v_2^\top \qone{1}}} & \leq \frac{(d-2)\cdot \exp(4)}{2} \cdot \exp\cp{\eta \cp{\frac{c_1''}{d^2}\sqrt{\frac{\log d}{m}} + \frac{c_1'}{d^2}\sqrt{\frac{\log d}{m}} - \frac{2d^{-2}}{\exp(4)}}} \\
        & \leq \frac{(d-2)\cdot \exp(4)}{2} \cdot \exp\cp{(c_1''+c_1')\cdot m\sqrt{\log d} - \frac{2}{\exp(4)}m^{3/2}} \\
        & \leq \frac{(d-2)\cdot \exp(4)}{2} \cdot \exp\cp{(c_1''+c_1')\cdot \cp{\log d}^4\sqrt{\log d} - \frac{2}{\exp(4)}\cp{\log d}^6} \\
        & \leq \frac{(d-2)\cdot \exp(4)}{2} \cdot \exp\cp{-c' \cp{\log d}^6} \\
        & \leq c\cdot d^{-c'}.
    \end{align*}
    Plug this back into Equation~\ref{proof_coro2_reciprocal}, there exists some constant \(C_1, C_1'>0\), such that for any \(C_1''>0\), with probability \(1-d^{-C_1}\), we have:
    \begin{equation}
        \atone[1]{1} + \atone[2]{1} \geq \frac{1}{1+C_1' \cdot d^{-C_1''}} \geq 1-C_1'\cdot d^{-C_1''}.
    \end{equation}
    For the second attention head, the proof is identical due to symmetry.
\section{Complete Proof of Lemma~\ref{lm2:symmetry_breaking}}\label{proof:lm2}
\begin{proof}
    W.l.o.g. we have assumed that the first query vector aligns slightly more with \(\mathbf v_1\) over \(\mathbf v_2\) at initialization. In Helper Lemma~\ref{helper_lemma:2}, we proved that, for any \(c>0\), with probability ar least \(1-m^{-c}\), we have \(\frac{\atone[1]{0}}{\atone[2]{0}} = \frac{\attwo[2]{0}}{\attwo[1]{0}} \geq 1+m^{-(0.5+c)}\) holds. Choosing \(c=0.5\) gives that:
    \begin{equation}\label{eq:lm2_initial_asym}
        \text{w.p. at least } 1-\frac{1}{\sqrt m}, \text{ we have }\frac{\atone[1]{0}}{\atone[2]{0}} = \frac{\attwo[2]{0}}{\attwo[1]{0}} \geq 1+\frac{1}{m}
    \end{equation}

    Then we will decompose the ratio as \(\frac{\atone[1]{1}}{\atone[2]{1}} = \frac{\atone[1]{0}}{\atone[2]{0}} \cdot \exp\cp{-\eta\left\langle\nabla_{\qone{0}}\mathcal L^{(0)}, \; \mathbf v_1 - \mathbf v_2\right\rangle}\). We will first lower bound the term \(-\left\langle \nabla_{\qone{0}} \mathcal L^{(0)}, \; \mathbf v_1 - \mathbf v_2\right\rangle\). Following the proof in Lemma~\ref{lm1:scale}, we consider the difference as the sum of four parts. For the first part, there exist constants \(c_1, c_1', c_1'' >0\), such that with probability greater than \(1-d^{-c_1}\):
    \begin{align*}
        \; T_1(1) - T_1(2) 
        = & \; \cp{2 \attwo[2]{0}\atone[1]{0}  - 2 \attwo[1]{0}\atone[2]{0}}\cdot (1-\delta_{12}) + 2 \attwo[1]{0}\atone[2]{0} \cp{\sum_{j\ne 2} \atone[j]{0} \delta_{2j}-\sum_{j\notin \{1,2\}}\atone[j]{0}\delta_{1j}} \\
        & \;  - 2 \attwo[2]{0}\atone[1]{0} \cp{\sum_{j\ne 1} \atone[j]{0} \delta_{1j} - \sum_{j\notin\{1,2\}}\atone[j]{0}\delta_{2j}} -2\attwo[1]{0}\atone[2]{0}\atone[1]{0} + 2\attwo[2]{0}\atone[1]{0}\atone[2]{0}\\
        \geq & \; \frac{c_1'}{d^2}\cp{\cp{\frac{\atone[1]{0}}{\atone[2]{0}}}^2-1} - c_1'' \cp{\frac{1}{d^2}\cdot \frac{\sqrt{\log d}}{m^{3/2}}} - \frac{2\exp(6)}{d^3}.
    \end{align*}
    Plug in the lower bound for \(\frac{\atone[1]{0}}{\atone[2]{0}}\) in statement~\ref{eq:lm2_initial_asym}, with probability at least \(1-\frac{1}{\sqrt m}-d^{-c_1}\), it follows:
    \begin{equation}\label{eq:lm2_t1}
        T_1(1) - T_1(2) \geq \frac{c_1'}{d^2m} - \frac{c_1''}{d^2}\frac{\sqrt{\log d}}{m^{3/2}} - \frac{2\exp(6)}{d^3}.
    \end{equation}
    For the second to the fourth term in Lemma~\ref{lm1:scale}, there exists constant \(c_2, c_2' >0\) such that with probability at least \(1-d^{-c_2}\):
    \begin{equation}\label{eq:lm2_t2_to_4}
        \sum_{j=2}^4 T_j(1) - T_j(2) \leq \sum_{j=2}^4 \cp{|T_j(1)| + |T_j(2)|} \leq c_2' \frac{\sqrt{\log d}}{d^2m^{3/2}}.
    \end{equation}

    Using union bound over Equations~\ref{eq:lm2_t1} and \ref{eq:lm2_t2_to_4}, there exist constants \(C_2, C_2', C_2'' >0\), such that with probability greater than \(1-m^{-C_2}\), we have:
    \begin{align*}
        \exp\cp{-\eta \left\langle\nabla_{\qone{0}}\mathcal L^{(0)}, \; \mathbf v_1 - \mathbf v_2\right\rangle} & = \exp\cp{\eta \sum_{i=1}^4 \cb{T_i(1) - T_i(2)}} \\
        & \geq \exp\cp{\eta \cp{T_1(1) - T_1(2)} - \eta\sum_{i=2}^4 |T_i(1) - T_i(2)|} \\
        & \geq \exp\cp{\eta \cp{\frac{C_2'}{d^2m} - C_2''\frac{\sqrt{\log d}}{d^2m^{3/2}}}}.
    \end{align*}

    Since we assume \qone{0} aligns more with \(\mathbf v_1\) than \(\mathbf v_2\) initially, it follows that there exist constants \(c_2, c_2', c_2''>0\) such that with probability greater than \(1-m^{-c_2}\):
    \begin{align*}
        \frac{\atone[1]{1}}{\atone[2]{1}} & = \exp\cp{\left\langle \qone{0}, \mathbf v_1 - \mathbf v_2\right\rangle - \eta\left\langle\nabla_{\qone{0}}\mathcal L^{(0)}, \; \mathbf v_1 - \mathbf v_2\right\rangle} \\
        & \geq \exp\cp{0- \eta\left\langle\nabla_{\qone{0}}\mathcal L^{(0)}, \; \mathbf v_1 - \mathbf v_2\right\rangle} \\
        & \geq \exp\cp{\eta \cp{\frac{c_2'}{d^2m} - c_2''\frac{\sqrt{\log d}}{d^2m^{3/2}}}}.
    \end{align*}
    Due to symmetry, we also have \(\frac{\attwo[2]{1}}{\attwo[1]{1}} \geq \exp\cp{\eta \cp{\frac{c_2'}{d^2m} - c_2''\frac{\sqrt{\log d}}{d^2m^{3/2}}}}\).
\end{proof}

\section{Proof of Corollary \ref{cor:sample-GD}}\label{proof:cor:sample-GD}
We begin with the observation that the loss gradient at initialization appears in Lemma \ref{lm1:scale} and the proof of Lemma \ref{lm2:symmetry_breaking} only through its interaction with the positional embeddings \(\left\langle\nabla_{\qone{0}}\mathcal L^{(0)}, \; \mathbf v_k\right\rangle\) or \(\left\langle\nabla_{\qone{0}}\mathcal L^{(0)}, \; \mathbf v_1 - \mathbf v_2\right\rangle\), and similar terms with respect to $\qtwo{0}$.
In the finite sample setting, we need to consider the empirical loss gradient $\nabla_{\qone{0}}\widehat{\mathcal L}^{(0)}$ instead of the population loss gradient $\nabla_{\qone{0}}\mathcal L^{(0)}$.
Since $\mathbf v_k$ are unit norm vectors, we have 
\[
\left| \left\langle\nabla_{\qone{0}}\widehat{\mathcal L}^{(0)}, \; \mathbf v_k\right\rangle - \left\langle\nabla_{\qone{0}}\mathcal L^{(0)}, \; \mathbf v_k\right\rangle \right| \leq \left\Vert \nabla_{\qone{0}}\widehat{\mathcal L}^{(0)} - \nabla_{\qone{0}}\mathcal L^{(0)}\right\Vert.
\]
Hence, the claims of Lemmas \ref{lm1:scale}--\ref{lm2:symmetry_breaking} hold with $\nabla_{\qone{0}}\widehat{\mathcal L}^{(0)}$ as long as the norm of the above deviation remains sufficiently small, specifically $O\left(d^{-2}\sqrt{\frac{\log d}{m}}\right)$, which is the noisy alignment with irrelevant positions stated in Lemma \ref{lm1:scale}.
We complete the proof by showing that for $n\geq md^4$ and $m \geq (\log d)^4$, with probability $1-m^{-c}$,
\[
\left\Vert \nabla_{\qone{0}}\widehat{\mathcal L}^{(0)} - \nabla_{\qone{0}}\mathcal L^{(0)}\right\Vert \leq \frac{C}{d^2}\sqrt{\frac{\log d}{m}}
\]
for some constants $c,C>0$. Naturally, the same arguments also hold for $\qtwo{0}$.
To derive the above tail bound, we first observe that for any \(\boldsymbol{\theta} \in\mathbb{R}^m\) and $\mathbf x \in \{\pm1\}^d$, $\ell(\mathbf x)= \cp{-x_1x_2+\cp{\sum_{i\in[d]} \gamma_i x_i}\cdot \cp{\sum_{i\in[d]} \tilde\gamma_i x_i}}^2 \in [0,4]$ implies that 
\begin{align*}
\left\Vert\nabla_{\boldsymbol{\theta}}\ell(\mathbf x)\right\Vert 
&= 2\sqrt{\ell(\mathbf x)}\cdot\left| \sum_{i=1}^d \tilde\gamma_i x_i\right| \cdot \left\Vert \sum_{i=1}^d x_i\nabla_{\boldsymbol{\theta}}\gamma_i\right\Vert
\\&= 2\sqrt{\ell(\mathbf x)}\cdot\left| \sum_{i=1}^d \tilde\gamma_i x_i\right| \cdot \left\Vert \sum_{i=1}^d \gamma_i (x_i\mathbf v_i) - \cp{\sum_{i=1}^d\gamma_ix_i}\sum_{j=1}^d \gamma_j \mathbf v_j\right\Vert
\quad\leq 8
\end{align*}
since $\{\gamma_i\}_{i\in[d]},\{\tilde\gamma_i\}_{i\in[d]}$ lead to convex combinations of $\{x_i\}_{i\in [d]} \in\{\pm1\}$ or unit norm vectors $\{\mathbf v_i\}_{i\in[d]},\{x_i\mathbf v_i\}_{i\in[d]}$.
As a consequence, by dominated convergence, $\nabla_{\qone{0}}{\mathcal L}^{(0)} = \nabla_{\qone{0}}\mathbb{E}_{\mathbf x}[\ell(\mathbf x)]= \mathbb{E}_{\mathbf x}[\nabla_{\qone{0}}\ell(\mathbf x)]$. We now observe that conditioned on $\qone{0}$, $\nabla_\theta \ell(\mathbf x_1),\ldots,\nabla_\theta \ell(\mathbf x_n)$ are independent. Hence, we can derive the tail bound using Hoeffding's inequality as
\begin{align*}
    &\mathbb{P}_{\qone{0},\{\mathbf x_s\}_{s\in[n]}}\cp{\left\Vert \nabla_{\qone{0}}\widehat{\mathcal L}^{(0)} - \nabla_{\qone{0}}\mathcal L^{(0)}\right\Vert > t}
    \\&=\mathbb{E}_{\qone{0}}\cb{\mathbb{P}_{\{\mathbf x_s\}_{s\in[n]}|\qone{0}}\cp{\left\Vert \sum_{s=1}^n \nabla_{\qone{0}}\ell(\mathbf x_s) - \mathbb{E}_{\mathbf x_s|\qone{0}} \cb{\nabla_{\qone{0}}\ell(\mathbf x_s)}\right\Vert > nt}}
    \leq 2\exp\cp{-\frac{nt^2}{8}} 
\end{align*}
For $t = \frac{C}{d^2}\sqrt{\frac{\log d}{m}}$ and $n\geq md^4$, the above bound is smaller than $d^{-C/8}$, which can be bounded by $m^{-c}$ for $m\leq d$. This completes the proof.
\section{Related work}\label{sec:related_works}
\paragraph{Feature Learning with Parities} While our work focuses on the canonical sparse XOR problem, the broader literature investigates the general parity problem \(k\geq 3\) to understand feature learning. Prior work establishes that sparse parity cleanly separates active feature learning of FFNNs from the lazy training dynamics of linear models in the NTK regime. Specifically, \citet{DanielyM20} proves an exponential separation in parameter efficiency between training a two-layer FFNN and a linear classifier over fixed embeddings. Beyond parameter efficiency, extensive research focuses on the sample or time complexity of learning parity functions and other sparse polynomials via stochastic gradient descent. \citet{AbbeAM23} provides a formal complexity bound by introducing the leap measure for sparse polynomials, proving that the high leap of \(k\)-parity problem places high sample and time requirements for FFNNs. To address these complexity bottlenecks, prior work explores curriculum learning for this task~\citep{AbbeCL23, CornacchiaM23}. In particular, \citet{CornacchiaM23} proves that, by carefully choosing the order of training examples, the required number of samples can be significantly reduced compared with training directly on samples from a uniform distribution.

\paragraph{Learning Dynamics of Transformers} Moving beyond FFNNs, recent work investigates how Transformers address parity learning. For instance, \citet{KimS25} analyzes how Transformers solve parities via Chain-of-Thought (CoT) reasoning, where the general parity task is sequentially decomposed into a series of intermediate XOR problems. However, their use of one-hot positional encodings results in a \(\Omega(d^2)\) parameter lower bound and is therefore inefficient. Besides parity, the broader literature extensively explores Transformer dynamics, ranging from empirical studies of deep, over-parameterized architectures~\citep{BhattamishraPKB23, LiuAGKZ23} to rigorous theoretical analysis from a mean-field perspective~\citep{Geshkovski23, Rigollet25}.

\paragraph{Performance Gap between Softmax Attention and Its Approximation}
Zooming in on the core mechanisms of the Transformer, a parallel line of work studies the performance gap in both expressivity and learnability between exact softmax attention and its linear or component-wise approximations. From the perspective of expressive power, \citet{HanPXHPLLS024} shows that linear attention can lead to semantic collapse by mapping distinct queries to an identical output. In contrast, exact softmax attention inherently prevents this bottleneck by guaranteeing an injective mapping. In addition, \citet{Xu26} demonstrates that linear variants are fundamentally insensitive to the norm of the queries, hence preventing the model from attending sharply to highly relevant signals. For learnability, \citet{Duranthon26} establishes a strict learning gap between softmax and linear attention on a single-location regression task, showing both theoretically and empirically that softmax can achieve the Bayes risk, while the linear substitute cannot.

Our work bridges different aspects of the literature by analyzing the learning dynamics of Transformers with the sparse XOR problem. First, we show that Transformers are more parameter-efficient at learning sparse Boolean functions than FFNNs. While our primary focus is to prove this parameter efficiency, we provide rigorous finite-sample bounds to guarantee generalization. Finally, to complement our theory, we empirically demonstrate that the one-step convergence dynamics depend on the softmax activation. Therefore, our work contributes as an empirical puzzle piece to the attention literature, showing that the necessity of softmax extends beyond the regression problem into the sparse classification task.

\end{document}